\documentclass[11pt,a4paper]{article}

\usepackage[T1]{fontenc}
\usepackage[utf8]{inputenc}
\usepackage{lmodern}
\usepackage[margin=2.5cm]{geometry}
\usepackage{amsmath,amssymb,amsthm}
\usepackage{graphicx}
\usepackage{booktabs}
\usepackage{multirow}
\usepackage{tabularx}
\usepackage{caption}
\usepackage{subcaption}
\usepackage{xcolor}
\usepackage{enumitem}
\usepackage{algorithm}
\usepackage{algpseudocode}
\usepackage[numbers,sort&compress]{natbib}
\usepackage[colorlinks=true,linkcolor=blue!60!black,citecolor=blue!50!black,urlcolor=blue!60!black]{hyperref}
\usepackage{doi}
\usepackage{microtype}

\newtheorem{definition}{Definition}
\newtheorem{proposition}{Proposition}

\graphicspath{{figures/}}

\title{Character-Name Concentration in LLM Fiction:\\Measuring Displacement and Mitigating Repetition}

\author{
Alejandro Mozo Quesada \\ Universidad de Málaga \\ \texttt{amozo@uma.es}
\and
Sergio Gálvez Rojas \\ Universidad de Málaga \\ \texttt{galvez@uma.es}
\and
Francisco López Valverde \\ Universidad de Málaga \\ \texttt{valverde@uma.es}
}

\date{}

\begin{document}
\maketitle


\begin{abstract}
Character names in large language model (LLM) fiction concentrate on a handful of recurrent names such as \emph{Elara} and \emph{Elias}, a measurable face of AI slop.
Existing fixes either require logit access or retraining, or operate per request and cannot control concentration across a corpus.
How much prompt-level strategies reduce concentration, and at what cost to name naturalness, has not been systematically measured.
We present an empirical study of name-concentration control across eight strategies and introduce speculative naming, a fine-tuning-free method that adapts the draft-verify-reject paradigm of speculative decoding from tokens to names.
The method requests candidate name-sets through verbalized sampling, scores them against a SlopTracker that combines a per-model fingerprint with a runtime counter, and selects through a temperature-controlled rule or its strict variant (HARD).
We use 20 LLMs to generate 44{,}782 stories in four languages from text and visual stimuli (88{,}919 named characters), and measure concentration with a list-free collision metric (Slop Ratio, SR) decomposed into known and emergent components.
List-based strategies do not remove slop: static banlists displace concentration to emergent names and more than halve the share of attested real names.
Speculative naming removes it: the sampling scaffold alone cuts SR by 89\% on average, the dynamic tracker brings the total reduction to 95\%, mean collision probability falls from 7.3\% to 0.4\%, and known-name collisions hold at zero while naturalness stays close to baseline (42--69\% attested by language) and within-story collisions drop from 4.3\% to 0.2\%.
The strategy ranking is stable across languages, model families, and name-counting conventions.
Name concentration is thus controllable at the corpus level without logit access or retraining; broader creative diversity beyond naming remains open.
\end{abstract}


\section{Introduction}
\label{sec:intro}

Creative writing is now one of the most common uses of large language models, and its output is fluent yet measurably less diverse than human writing: major providers converge on the same idiosyncratic phrases regardless of architecture or training data~\citep{Chakrabarty2025salvaged, Guo2025benchmarking, Reinhart2025llms}, and the alignment processes that make models helpful reduce an automated index of linguistic creativity by about 30\%~\citep{Lu2024salieri}.
The deficit operates as a social dilemma: any single user benefits, while the aggregate effect is a narrowing of the creative landscape~\citep{Doshi2024generative}.
The plot-diversity gap between model output and human writing is 7 to 9 times~\citep{Xu2025echoes}, it widens as more text is generated~\citep{Moon2025homogenizing}, and synthetic text fed back as training data closes the loop~\citep{Sourati2026homogenizing}.

AI slop manifests at many levels of text.
The same aligned models repeat sensory cliches such as ``a shimmer of light'', intensifiers such as ``profoundly'', formulaic constructions such as ``It's not X, it's Y'', and character names at rates far above human text~\citep{Reinhart2025llms, Paech2025antislop}.
Among these units, character names stand out as an extreme and easily measured case, so this paper treats them as its case study.
Certain names appear orders of magnitude more frequently in model output than in human text.
Across thousands of stories from several providers, the large majority contain one of a handful of recurrent names~\citep{Paech2025antislop, Hamilton2026elias}.
This ``name slop'' is systematic.
The same attractor names recur across models, languages, and prompt variations, which points to a structural bias in how models generate creative choices rather than occasional lapses.

The root causes of this mode collapse have been traced to the alignment process itself.
Supervised fine-tuning reduces token-level diversity, while preference optimization methods such as Direct Preference Optimization (DPO) and Reinforcement Learning from Human Feedback (RLHF) cause a large drop in output diversity~\citep{Kirk2024understanding, Winata2025preference, Liu2026alignment}.
Human annotators show a \emph{typicality bias}, preferring familiar content, which compresses the learned distribution toward statistically average outputs~\citep{Zhang2025verbalized}.
Proposed solutions span modified preference objectives, entropy-controlled decoding, inference-time guidance, and prompting strategies.
However, the most effective approaches (logit-level sampling and model retraining) require access that most users lack.

To our knowledge, no prior work has systematically compared name-slop mitigation strategies that operate exclusively at the prompt and generation-architecture level, without logit access or fine-tuning.
Existing evaluations study a single strategy in isolation or measure general lexical diversity rather than the specific, high-concentration phenomenon of name slop.
We address this gap by evaluating eight configurations across 20 models, including a baseline, an anti-slop prompt, a banlist, two-pass generation, a multiagent reviewer, and three variants of speculative naming.

Our research questions are:
\begin{itemize}[noitemsep]
  \item \textbf{RQ1.} To what extent does each strategy reduce name-slop concentration?
  \item \textbf{RQ2.} Does name slop differ between text and vision input modes?
  \item \textbf{RQ3.} Is the multiagent approach more effective than prompt engineering, and does it justify its cost?
  \item \textbf{RQ4.} Does two-pass generation, which separates naming from content, remove known slop, or does concentration re-emerge on new names?
  \item \textbf{RQ5.} Do strategies perform equally across four languages, or are slop names language-specific?
  \item \textbf{RQ6.} Does enabling reasoning mode change name diversity?
\end{itemize}

Our contributions are:
\begin{enumerate}[noitemsep]
  \item \textbf{Speculative naming}, a fine-tuning-free, logit-free method that applies draft-verify-reject at the name level through verbalized sampling and a SlopTracker. It works with any model accessible through an API.
  \item \textbf{The Slop Ratio with known and emergent decomposition}, a list-free collision metric that separates removal from displacement. It exposes the balloon effect of banlist strategies: banning known names lowers the known component but raises the emergent one.
  \item \textbf{A large-scale empirical evaluation} across 20 LLMs, four languages, eight strategies, and 44{,}782 stories, with bootstrap CIs, effect sizes, and Holm-corrected tests. Dynamic runtime scoring yields the lowest concentration at a fixed cost premium (RQ1, RQ3); banning known names displaces concentration onto new names (RQ4); vision input lowers slop in four of five dual-mode families (RQ2); slop profiles are model- and language-specific, and reasoning mode leaves them unchanged (RQ5, RQ6). Against civil-registry name lists, speculative naming preserves naturalness close to baseline whereas banlists push toward invented names; the ranking is stable across languages, families, and counting conventions.
\end{enumerate}


\section{Background and Related Work}
\label{sec:related}

\subsection{Defining and measuring slop}

\citet{Paech2025antislop} defined ``slop'' as phraseology that appears orders of magnitude more frequently in model output than in human text, and sorted it into four types: character names, sensory clich\'es, intensifiers, and formulaic constructions.
\citet{Shaib2025measuring} gave slop a formal taxonomy, with three themes and seven codes, and found that language models cannot reliably detect it.
Each model has a distinct slop fingerprint, such as \emph{Elara} for Gemma or \emph{Elias} for other models~\citep{Hamilton2026elias, Rudko2025chatgpt}.
Complementary lines of evidence quantify the human--model gap at other linguistic levels, from lexical diversity~\citep{Kendro2025lexical} to syntactic flatness under formal grammar~\citep{Zamaraeva2025hpsg}, and the story-evaluation literature catalogs the available metrics~\citep{Yang2024survey, Teleki2025survey}, including reference-based creativity scoring~\citep{Li2025ttcw}.

\subsection{Causes: typicity bias and the alignment tax}

\citet{Zhang2025verbalized} identified typicity bias as a root cause of mode collapse.
Human annotators prefer familiar content, which compresses the distribution toward safe modes.
\citet{Kirk2024understanding} found that reinforcement learning from human feedback reduces the diversity of model outputs relative to supervised fine-tuning. \citet{Liu2026alignment} traced the collapse to the preference-optimization stage: supervised fine-tuning preserves diversity, DPO drives it down, and 40 to 79\% of questions in TruthfulQA produce a single semantic cluster after alignment, which shows that the collapse is intrinsic to the learned distribution rather than to the decoding step.
Even chat templates cause systematic diversity collapse~\citep{Yun2025price}.
\citet{Hamilton2026elias} traced the ``Elias/lighthouse'' slop pattern to 3{,}053 stories in the post-training data of one open model, an end-to-end attribution of a slop pattern to its source training documents.
The consequences show up wherever model output accumulates: group ideation produces more homogeneous ideas~\citep{Anderson2024homogenization}, model panels converge on open-ended questions~\citep{Jiang2025hivemind}, LLMs trail humans on divergent thinking~\citep{BellemarePepin2026divergent}, and aggregate creativity indices have stopped improving~\citep{Haase2025peaked}; originality differences owe more to prompts than to models~\citep{Haase2026within}, and the size of the human--model gap depends on the task~\citep{Yang2026asymmetry}; narration-based probes detect mode collapse without logit access~\citep{Hamilton2024detecting}, which supports the feasibility of the text-level control we pursue here.

\subsection{Mitigation: fine-tuning and decoding}

The most direct fix is to change what the model prefers before generation starts.
\citet{Paech2025antislop} proposed Final Token Preference Optimization (FTPO), which suppresses slop tokens with about 90\% effectiveness and under 1\% quality loss.
\citet{Lanchantin2025divpo} introduced Diverse Preference Optimization (DivPO), which selects preference pairs by rarity and quality; \citet{Deshpande2025diverse} showed that common diversity metrics are biased toward short text and proposed a length-controlled alternative; \citet{Ismayilzada2025crepo} injected novelty and surprise signals into DPO; later work regularizes the preference objective itself, decoupling diversity from likelihood loss~\citep{Slocum2025spl}, trading diversity against a quality floor~\citep{Pan2026qempo}, and countering preference over-optimization~\citep{Zhai2026uprlhf}.
The price is access: weights and a retraining cycle for every model version.

Decoding strategies need no training, but they operate at the token level and require logit access.
Min-$p$ sampling scales the truncation threshold by the top token probability~\citep{Nguyen2025minp}; Top-H decoding formalizes the creativity--coherence trade-off as bounded-entropy optimization~\citep{Baghaei2025toph}; DiffSampling truncates by a discrete derivative of the probability curve~\citep{Franceschelli2025diffsampling}; geometric regulation reshapes the value cache to escape collapse~\citep{Du2026geometric}; G2 adds diversity and deduplication guides during decoding~\citep{Ruan2025g2}; Avoidance Decoding penalizes similarity to previously generated outputs~\citep{Park2025avoidance}.
The closest dynamic relative is the antislop sampler distributed with \citet{Paech2025antislop}, which blocks and resamples slop phrases across outputs at the logit level; it presupposes token-probability access, which speculative naming does not.
Prompt-level methods need neither weights nor logits: Verbalized Sampling asks the model to state a distribution over responses with probabilities, which restores diversity~\citep{Zhang2025verbalized}; guided prompting with explicit diversity constraints helps~\citep{Ruan2025g2}, whereas fine-grained persona prompting alone does not raise lexical diversity~\citep{Kambhatla2025persona}.

\subsection{Multi-agent and multi-pass generation}

A second family orchestrates several model calls around a single generation: multi-agent debate improves factuality and reasoning~\citep{Du2024improving}, collaborative critique adapts to story generation~\citep{Bae2024collective} though debate can amplify bias~\citep{Ma2025judging}, and multi-pass pipelines alternate planning, writing, and revision~\citep{Chi2025thoughtsculpt, Wu2025superwriter, Wan2025cogwriter}; multi-LLM collaboration distributes authorship with little quality loss but reduced lexical diversity~\citep{Venkatraman2025collabstory}.
These methods improve planning and coherence at a multiplied call cost, and they act on the response as a whole: none of their monitors scores a repeated character name.

\subsection{Summary and research gap}

Each level leaves the name-slop problem open.
Fine-tuning needs weights and retraining, and its effect is model-wide rather than targeted at one slop unit.
Decoding methods require logit access and cannot tell that a character name repeats across stories.
Prompt-level methods restore response diversity without special access, but neither identify nor control a specific slop unit; multi-agent and multi-pass methods improve planning and revision with generic critique criteria, and none of their monitors scores name repetition.
Prompt-level diversity methods, including verbalized sampling, share a deeper structural limit: they are memoryless and per-request, while name slop is corpus-level and accumulates across requests.
The gap is precise: no existing method treats a character name as a distinct control unit with its own draft-verify-reject loop, available without logit access or retraining; a name is a single decision that propagates through an entire story, which is what makes it unusually tractable to control.


\section{The Proposed Method}
\label{sec:method}

This section gives the formal definition of speculative naming.
We state the problem, define the SlopTracker, derive the selection rule, present the algorithm, and state its properties.

\subsection{Problem formulation}

We generate stories under a fixed prompt with placeholder slots for character names.
Let a \emph{scaffold} be a story text with $k$ placeholders, such as \texttt{[CHAR\_1]}, \ldots, \texttt{[CHAR\_$k$]}.
The goal is to fill each placeholder with a name such that (a)~the name is coherent with the story and (b)~the name does not repeat across stories beyond a tolerable frequency.
The first condition is aesthetic and is judged by the model.
The second condition is empirical and is judged by observed data.
The formal objects that follow are defined over any countable unit an extractor can isolate; this paper instantiates them with character names, and Section~\ref{sec:beyond-names} returns to other units.

The name echoes speculative decoding~\citep{Leviathan2023fast, Chen2023speculative}, which drafts tokens with a small model and verifies them against a target model.
Two substitutions adapt the paradigm.
The unit is a name rather than a token, and the verifier is an empirical anti-repetition constraint rather than a reference model; no logits are inspected, which is what makes the method usable with any model accessible through an API.

\subsection{The SlopTracker}

The SlopTracker maintains two frequency maps for each model $m$ and each experimental cell $c$.

\begin{definition}[Baseline fingerprint]\label{def:fingerprint}
The baseline fingerprint $B_m$ is a frequency map built once from the model's baseline output and frozen before any strategy run:
$$B_m(n) = \frac{\#\text{characters named } n \text{ in baseline stories of model } m}{\text{total named characters in baseline stories of model } m}.$$
It assigns a value to every name observed in baseline output, pooled over languages and input modalities.
Names absent from baseline output have $B_m(n) = 0$.
\end{definition}

\begin{definition}[Runtime counter]\label{def:runtime}
The runtime counter $R_c(t)$ is a frequency map updated during generation, over the first $t$ stories of cell~$c$, initialized at $R_c^{(0)}(n) = 0$ for every $n$:
$$R_c(t)(n) = \frac{\#\text{characters named } n \text{ among the first } t \text{ stories}}{\text{total named characters among the first } t \text{ stories}}.$$
The runtime component stays inactive until the counter has accumulated more than $N_{\min} = 50$ named characters (cold start); before that threshold, $R_c(t)(n) = 0$ for all $n$.
\end{definition}

\begin{definition}[Slop score]\label{def:slopscore}
The slop score of a candidate name $n$ at step $t$ combines the two sources of evidence by taking their maximum:
$$s(n) = \max\bigl(B_m(n),\; R_c(t)(n)\bigr).$$
\end{definition}

The maximum is chosen so that a name is penalized if it concentrates either in the model's prior output or in the current run.
The baseline fingerprint catches known slop from the start.
The runtime counter catches emergent slop that was absent from baseline but accumulates during generation.
The counter is indexed by character entities, not textual mentions: a placeholder receives a single name that is substituted at every occurrence, so repeating the same character within a story adds one entry, while two distinct characters sharing a name add two, which is precisely the collision SR measures.

Three scope conventions complete the definition.
First, a cell $c$ is a model--strategy pair whose counter is shared across the four generation languages and both modalities: a name that accumulates in English raises the penalty for the same name in Spanish within the cell (deliberate: the target of control is corpus-level concentration).
Second, names are normalized before scoring: first given name, lowercased and stripped of diacritics; single-character tokens are discarded.
Third, the tracker scores each name in a candidate set independently and does not prevent two placeholders from receiving names with the same normalized form; such within-set collisions are recorded after selection and penalized later, but not prevented inside the story where they occur (Section~\ref{sec:limitations}).

\subsection{The selection rule}

The model produces $K$ candidate name-sets, each with a self-reported probability $\hat{p}_i \in [0,1]$ that sums to 1 (verbalized sampling).
Candidate set $i$ is selected with probability
$$P(i) = \frac{\hat{p}_i \cdot \exp\bigl(-\lambda \cdot s_i\bigr)}{\sum_{j=1}^{K} \hat{p}_j \cdot \exp\bigl(-\lambda \cdot s_j\bigr)},$$
where $s_i = \sum_{n \in \text{set}_i} s(n)$ is the total slop score of the set, and $\lambda \geq 0$ is the anti-slop parameter.

This rule is a Boltzmann reweighting.
The numerator combines two signals multiplicatively.
The first factor, $\hat{p}_i$, is the model's stated coherence: how fitting the model finds the name.
The second factor, $\exp(-\lambda s_i)$, is a freshness term that decays with observed concentration.
The parameter $\lambda$ acts as an inverse temperature.
When $\lambda = 0$, the rule reduces to pure verbalized sampling.
As $\lambda$ grows, the distribution shifts toward low-slop names.
A name can be both highly coherent and highly slop (\emph{Elara} is a model's preferred name and its most frequent one).
The product combines the two rather than choosing one.

\medskip\noindent\textbf{Property (soft-rule limit).}
Let $I_* = \{i : s_i = \min_j s_j\}$ be the candidates tied at the minimum total score.
If all stated probabilities are positive, then
$$\lim_{\lambda \to \infty} P(i) =
\begin{cases}
\hat{p}_i \big/ \sum_{j \in I_*} \hat{p}_j, & i \in I_*, \\
0, & i \notin I_*.
\end{cases}$$
The limit restricts sampling to the lowest-scoring candidates and redistributes their stated probabilities proportionally.
It does not reduce to a deterministic choice of the most probable candidate: two tied candidates with stated probabilities $0.6$ and $0.4$ keep those probabilities in the limit.

\medskip\noindent\textbf{Definition (HARD mode).}
HARD is a strict-threshold variant, not the limit of the soft rule.
A candidate set is \emph{admissible} when every name in it has slop score $s(n) < \tau$, with $\tau = 0.02$.
HARD sorts candidate sets by stated probability in decreasing order and selects the first admissible set.
If no set is admissible, it falls back to the candidate with minimum total score $s_i$.
Both rules share the same preference for low-score candidates; HARD replaces smooth reweighting with a hard admissibility cutoff and a deterministic choice among admissible sets.

\subsection{Algorithm}

Algorithm~\ref{alg:spec} assembles the pieces into a single generation loop.
It keeps the two-pass scaffold intact and inserts a scoring and selection layer between the two passes, which adds one model call over plain two-pass generation.

\begin{algorithm}[t]
\caption{Speculative naming}\label{alg:spec}
\begin{algorithmic}[1]
\Require model $m$, prompt scaffold, baseline fingerprint $B_m$, parameter $\lambda$
\Ensure a named story
\State \textbf{SCAFFOLD.} Generate a story with placeholders \texttt{[CHAR\_1]}, \ldots, \texttt{[CHAR\_$k$]}.
\State \textbf{DRAFT.} Request $K$ candidate name-sets, each with verbalized probability $\hat{p}_i$.
\State \textbf{VERIFY.} For each name $n$ in candidate set $i$, compute $s(n) = \max(B_m(n), R(n))$ and $s_i = \sum_n s(n)$.
\State \textbf{WEIGHT.} Compute $w_i = \hat{p}_i \cdot \exp(-\lambda \cdot s_i)$.
\State \textbf{SELECT.} Sample set $i$ with probability $w_i / \sum_j w_j$.
\Statex \hspace{2em}(HARD mode: among sets whose names all have $s(n) < 0.02$, select the one
\Statex \hspace{2em}with maximum $\hat{p}_i$; if none is admissible, select minimum $s_i$.)
\State \textbf{SUBSTITUTE.} Replace placeholders with the selected names.
\State \textbf{UPDATE.} Add the selected names to the runtime counter $R$.
\end{algorithmic}
\end{algorithm}

\begin{figure}[t]
  \centering
  \includegraphics[width=\textwidth]{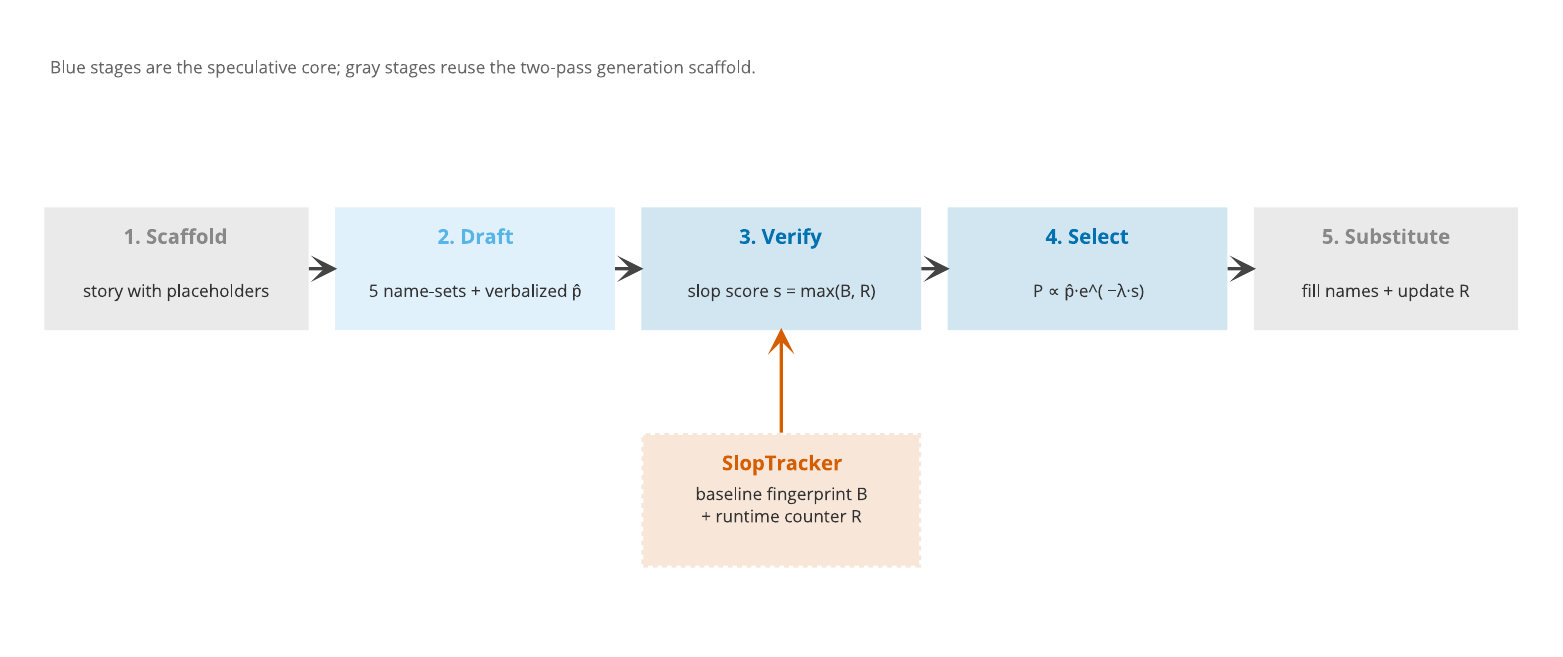}
  \caption{The speculative naming pipeline. The model first writes a story with placeholders (scaffold), then drafts five name-sets with verbalized probabilities (draft). The SlopTracker, which combines a baseline fingerprint with a runtime counter, scores each candidate by its observed frequency (verify). Selection weights each candidate by the product of its stated probability and a freshness term (select), the placeholders are filled, and the tracker is updated (substitute). Blue stages are the speculative core; gray stages reuse the existing two-pass scaffold. The whole loop adds one extra model call over plain two-pass generation.}
  \label{fig:method_schema}
\end{figure}

Steps~1 and~6 are plain two-pass generation: the model writes the story with placeholders and fills them afterwards.
Step~2 is the drafting call: it requests $K$ candidate name-sets, each with a verbalized probability.
Steps~3 to~5 are the speculative core.
The \textsc{verify} step scores each name by the larger of its baseline fingerprint frequency and its runtime frequency within the current run, and scores a candidate set by the total of its names' scores.
The \textsc{weight} step multiplies each stated probability by the freshness term, and the \textsc{select} step samples accordingly; \textsc{hard} mode, the unnumbered lines, applies the strict-threshold rule instead.
Step~7 closes the loop: the selected names enter the runtime counter, so later stories in the same run penalize them a little more.
Figure~\ref{fig:method_schema} traces the full pipeline; the blue stages are the speculative core and the gray stages reuse the two-pass scaffold.

\section{The Proposed Metric}
\label{sec:slopratio}

This section defines our evaluation metric, states its properties, and explains why it is list-free.

\subsection{Definition}

Let a cell contain $N$ named-character occurrences, with name $i$ appearing $c_i$ times ($\sum_i c_i = N$).
Let $V$ be the number of unique names.

\begin{definition}[Slop Ratio]\label{def:sr}
The Slop Ratio of the cell is
$$\mathrm{SR} = \frac{\sum_{i=1}^{V} c_i(c_i - 1)}{N(N-1)}.$$
SR is defined for $N \geq 2$; a cell with fewer than two named characters cannot exhibit a collision and is assigned $\mathrm{SR} = 0$ by convention.
\end{definition}

\subsection{Properties}

\begin{proposition}[Collision probability]\label{prop:collision}
SR equals the probability that two characters drawn uniformly at random without replacement from the cell share the same name. It is Simpson's concentration index with finite-sample correction.
\end{proposition}
\begin{proof}
$\mathrm{P}(\text{same}) = \sum_i \mathrm{P}(\text{first}=i)\cdot\mathrm{P}(\text{second}=i \mid \text{first}=i) = \sum_i (c_i/N)\cdot((c_i-1)/(N-1)) = \mathrm{SR}.$ \qed
\end{proof}

\begin{proposition}[Bounds]\label{prop:bounds}
$\mathrm{SR} \in [0,1].$ Further, $\mathrm{SR}=0$ if and only if every name is distinct, and $\mathrm{SR}=1$ if and only if all characters share one name. For a fixed number of unique names $V$, SR is minimized when the distribution is uniform.
\end{proposition}

\begin{proposition}[Additive decomposition]\label{prop:decomp}
Let $K$ be the set of names whose baseline frequency reaches 2\%, $K = \{n : B_m(n) \geq 0.02\}$; a concentration in $K$ counts as known slop, and any other name counts as emergent slop.
Define
$$\mathrm{SR}_{\text{known}} = \frac{\sum_{i \in K} c_i(c_i-1)}{N(N-1)}, \qquad \mathrm{SR}_{\text{emergent}} = \frac{\sum_{i \notin K} c_i(c_i-1)}{N(N-1)}.$$
Then $\mathrm{SR} = \mathrm{SR}_{\text{known}} + \mathrm{SR}_{\text{emergent}}.$
\end{proposition}
\begin{proof}
The index set $\{1,\ldots,V\}$ partitions into $K$ and its complement. The sum in Definition~\ref{def:sr} splits accordingly, and both terms are non-negative. \qed
\end{proof}

The decomposition is the analytic core of this paper.
A strategy that \emph{eliminates} slop drives both components toward zero.
A strategy that \emph{displaces} slop drives the known component toward zero while leaving the emergent component high, because banned names are replaced by new names that then concentrate.
This distinction is invisible to any metric that measures concentration against a fixed banlist, because the new names are not on the list.

\subsection{Why the metric is list-free}
\label{sec:listfree}

SR depends only on the observed name frequencies within a cell and refers to no external slop vocabulary.
This matters because slop names are largely language-specific (mean cross-language Jaccard overlap of the top 50 names at baseline: 0.094; \emph{Elara}, \emph{Elias}, and \emph{Leo} recur in all four languages, but no attractor is shared everywhere), so a monolingual banlist does not transfer, whereas SR is computed identically in every language and every model.

We also report four conventional diversity metrics: top-5 concentration $C_5$, Shannon entropy $H$, the number of unique names $V$, and Pielou evenness $E = H/\log_2 V$, with the convention $E = 0$ when $V \leq 1$ (a single unique name is maximal concentration, not perfect evenness).
All slop metrics are computed on proper-name characters only.
We exclude unnamed or role-based characters (such as ``the detective'') and stimulus words used as names (such as ``Phoenix''), because these are not creative naming choices.
One model (Llama~3.3~70B) barely names characters at baseline: proper names appear in only 12\% of its stories, and about one in five of its characters is a stimulus word used as a name. Its baseline SR rests on roughly 30 characters, so we exclude it from the descriptive cross-model averages of Section~\ref{sec:decomposition} (19 models). We retain it in the per-model paired tests of Section~\ref{sec:stats}, where each model serves as its own control.


\section{Experimental Setup}
\label{sec:setup}

\subsection{Corpus}

We generated 44{,}782 stories across 20 models: 160 experimental cells (one per model--strategy pair over the eight strategies of Section~\ref{sec:strategies}), plus supplementary runs (a seven-model subset with intermediate speculative penalties, min-p variants, and no-stimulus conditions for the 15 chat models).
Each cell combines one model, one strategy, one input modality (text or vision), and one reasoning setting.
We extracted 88{,}919 named characters with a fixed extractor model, validated as equivalent to a second extractor on 20 control stories (100\% match): 57{,}878 proper names, 27{,}554 unnamed or role-based, 3{,}487 stimulus words used as names.

\subsection{Models}
\label{sec:models}

The 20 models span five architecture families and a range of sizes: Gemma~3~27B (chat, vision, and an FTPO-trained antislop variant), Qwen~3.5~9B (chat, vision), Qwen~3.6~35B (chat, vision, thinking), GPT-OSS~20B (chat, thinking), GPT-5.4~Nano, DeepSeek~V4~Flash (chat, thinking), Ministral~3~14B (chat, vision), Nemotron~3~Nano~Omni (chat, thinking), Llama~3.3~70B, and Gemini~3.1~Flash~Lite, a mix of open and proprietary, base and instruction-tuned, text-only and vision-capable systems.

\subsection{Strategies}
\label{sec:strategies}

We compare eight strategies.
The first five are standard baselines.
The last three are the speculative method at three parameter settings.

\begin{enumerate}[noitemsep]
  \item \textbf{Baseline.} The model generates the story with names as it sees fit.
  \item \textbf{Anti-slop prompt.} The model is instructed to avoid clich\'ed names.
  \item \textbf{Ban top 20.} The model receives an explicit list of its 20 most frequent baseline names and is told not to use them.
  \item \textbf{Multiphase.} The model writes the story with placeholders, then assigns names in a second pass.
  \item \textbf{Multiagent.} A reviewer extracts names, queries a dynamic database, and requests reformulation if any name exceeds three uses per model or 1\% of all stories. Up to three revision rounds.
  \item \textbf{Speculative $\lambda=0$.} Verbalized sampling with placeholder substitution and no penalty: the memoryless limit of the method and the neutral control for the scaffold.
  \item \textbf{Speculative $\lambda=10$.} Moderate anti-slop penalty.
  \item \textbf{Speculative HARD.} Reject any candidate name whose slop score reaches the threshold ($s(n) \geq \tau = 0.02$).
\end{enumerate}

\subsection{Stimulus and languages}

The stimulus is a set of five pictograms presented as an image (vision mode) or a textual list (text mode), all five to be incorporated in the given order, which controls story structure across conditions.
Stories are generated in four languages: English (40\% of stories per cell), Spanish, French, and German (20\% each); multilingual creativity assessment is an active area~\citep{Luchini2025multilingual}, yet mitigation studies rarely test language transfer.
Name scoring shares its state across languages (fingerprint pooled over baseline output, runtime counter over all four languages), so the concentration being controlled is corpus-level (Section~\ref{sec:method}).

\subsection{Metrics}

We compute five diversity metrics per cell: Slop Ratio (SR), top-5 concentration ($C_5$, lower is better), Shannon entropy ($H$, higher is better), unique names ($V$, higher is better), and evenness ($E$, higher is better).
We also report the SR decomposition into known and emergent components.
To check that name diversification does not degrade the prose, we measure lexical diversity (Distinct-2, PATTR~\citep{Deshpande2025penalty}, and MATTR-50), inter-story semantic similarity (mean pairwise cosine over MiniLM embeddings), and cross-language name overlap (Jaccard).

\subsection{Statistical analysis}
\label{sec:stats}

We report bootstrap 95\% CIs (1{,}000 iterations), Cohen's $d$, and Wilcoxon signed-rank tests across the 20 models with Holm--Bonferroni correction; paired tests include all 20 models (each model is its own control), while the descriptive averages of Section~\ref{sec:decomposition} use 19 (Section~\ref{sec:listfree}).
A strategy ``wins'' a metric for a model if it achieves the best value among the eight strategies.

\subsection{Cost}

Speculative naming uses two model calls and costs 3.30 times the baseline in tokens, because the second call generates five name-candidate sets.
The cost is fixed, unlike multiagent, whose cost varies with the number of revision rounds, from 470 to 16{,}885 tokens per story (2.65 calls on average).
Ban top 20 is not free either: the banlist itself inflates the prompt by about a third.
Multiphase is the cheapest two-pass option at 2.99 times baseline.
The full per-strategy costs appear with the cost-effectiveness analysis (Table~\ref{tab:cost_effectiveness}, Appendix~\ref{app:A}).


\section{Results}
\label{sec:results}

\subsection{Global comparison: speculative HARD yields the lowest concentration (RQ1)}

Table~\ref{tab:wins} (appendix) shows how many of the 20 models each strategy wins on each metric (100 comparisons total).
Speculative HARD wins 57 of 100, more than the next three strategies combined ($18 + 15 + 10 = 43$).
It ranks best by average rank in 15 of 20 models.
The ranking is not driven by the low-slop outlier: excluding Llama~3.3~70B, whose baseline rests on roughly 30 named characters, speculative HARD still wins 53 of 95 comparisons and ranks best in 14 of 19 models.
Speculative $\lambda=10$ ranks best in the remaining five models, four of which are thinking variants (the fifth is GPT-OSS:20b).
No standard strategy wins any model on the overall average rank.

Once the speculative method is included, no standard strategy wins on any slop metric (SR or $C_5$).
Standard strategies are dimension-specialists.
Multiphase wins unique names in 11 models, because its placeholder substitution forces proper naming and produces the most named characters per story.
But it does not reduce slop, because the name-assignment step has no frequency control.
Speculative HARD wins on all five metrics and combines the strengths of both: it forces proper naming (like multiphase) and reaches the lowest SR we observe (lower than multiagent).

Speculative HARD reduces SR by 95\% on average; the full significance picture is in the forest plot of Figure~\ref{fig:forest_plot}.

\subsection{The SR decomposition reveals the balloon effect}
\label{sec:decomposition}

\begin{figure}[t]
  \centering
  \includegraphics[width=0.85\textwidth]{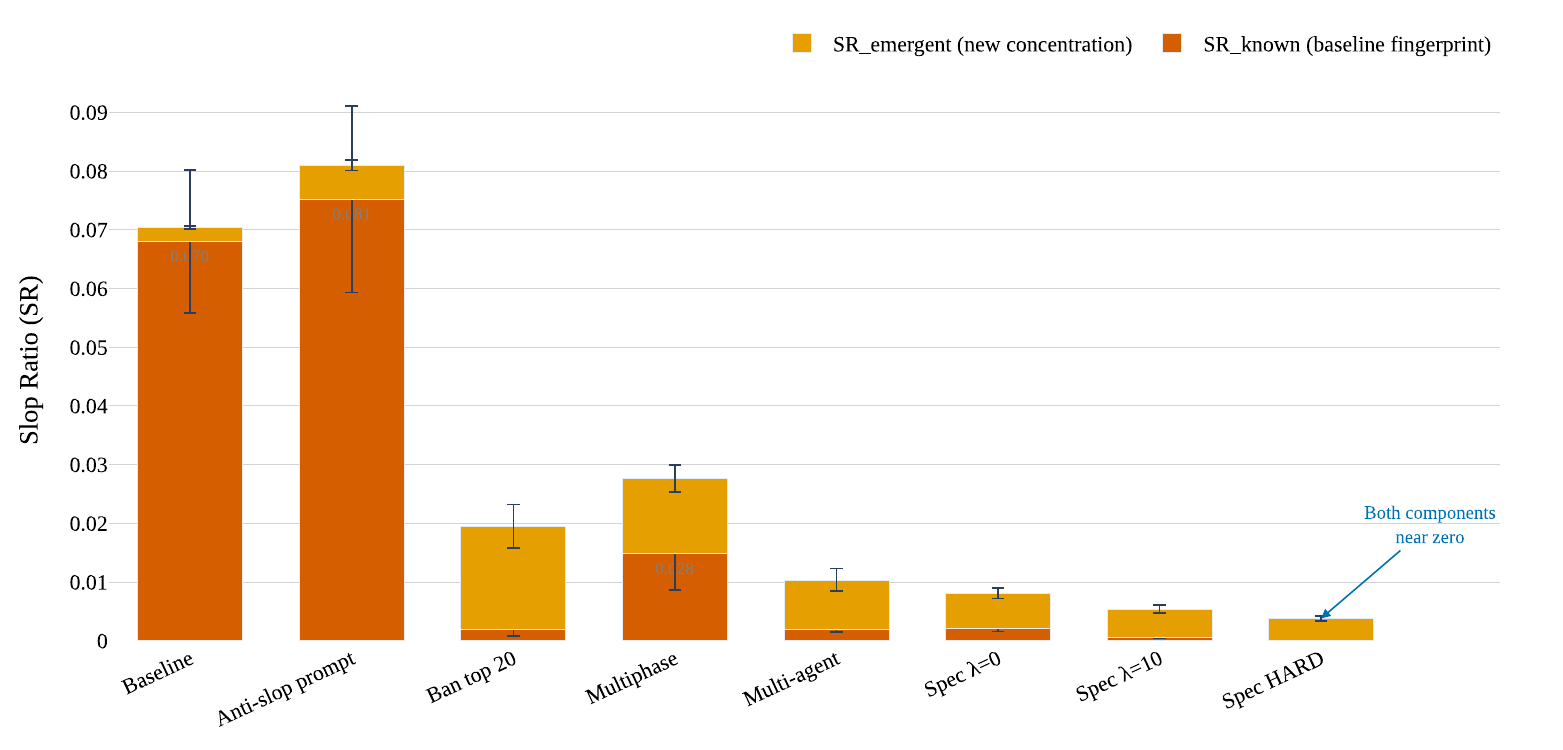}
  \caption{Slop Ratio decomposition by strategy. Mean SR across 19 models, stacked into the known component (concentration in baseline-fingerprint names, dark red) and the emergent component (concentration in new names, orange). Error bars show one standard error. Speculative HARD (rightmost, annotated) is the only strategy that brings both components near zero. Ban top 20 drives the known component to zero but shows the highest emergent component, the balloon effect. The anti-slop prompt leaves both components near baseline.}
  \label{fig:sr_decomposition}
\end{figure}

Figure~\ref{fig:sr_decomposition} shows the SR decomposition for each strategy, averaged across the 19 models with comparable naming behavior (Llama~3.3~70B excluded).

Ban top 20 drives the known component of SR close to zero (0.0021 versus a baseline of 0.0703), but the emergent component is the highest of any strategy (0.0142).
Forbidden from its familiar names, the model generates new ones that concentrate just as heavily: \emph{Hemlock} appears 44 times for Gemma, and \emph{Fitzwilliam} becomes the top name for the FTPO antislop variant (51 occurrences).
This is the balloon effect.

Multiagent reduces both components (known 0.0017, emergent 0.0082).
Its fixed threshold of three uses per name prevents any name from concentrating, whether slop or new.
Speculative HARD is the only strategy that brings both components near zero (known 0.0000, emergent 0.0036).
The zero known component is expected by construction: the threshold rule rejects candidates whose names carry a known-slop score, so a low known SR alone is weak evidence for any rejection-based method.
Two observations carry the weight instead.
First, the emergent component, which no other strategy brings below 0.008, shows that HARD does not merely displace concentration onto new names.
Second, the reduction does not come at the cost of naturalness: against civil-registry name lists, HARD keeps a share of attested real names close to baseline, whereas the banlist, whose known component is also near zero, does not (Section~\ref{sec:naturalness}).
The dynamic SlopTracker achieves this because it accumulates runtime frequencies.
Names that emerge as new attractors during generation are progressively penalized, which prevents the second-layer concentration that plagues static banlists.

Fine-tuning shows the same pattern: the FTPO model of \citet{Paech2025antislop} suppresses its banlist names while new attractors grow (Section~\ref{sec:discussion}).

\subsection{Per-model analysis generalizes across architectures}

The speculative HARD effect is consistent across families, sizes, and modalities: from the deepest-slop cells (Qwen 3.6 Thinking: 0.177 to 0.002) through moderate ones (Gemma 3 27B: 0.091 to 0.004), and across all four online models (Gemini Flash Lite 0.188 to 0.006, GPT-5.4 Nano 0.038 to 0.001, both 97\% reductions).
No model family resists the method, which is why the global win count is led by speculative HARD rather than by any architecture-specific interaction.

Naming behavior itself varies widely: the share of proper names among named characters ranges from 17\% (Llama 3.3 70B, which prefers role descriptions and stimulus words) to 93\% (Gemma Antislop).
The deep-slop models are proper-name-heavy, which is where slop has room to concentrate.
Two-pass strategies force naming, and when name-averse models are forced to name, they produce extreme concentration: under multiphase, Llama's proper-name share rises from 17\% to 91\% and it assigns \emph{Kaid\"{e}n} 61 times.

\subsection{Name naturalness and robustness checks}\label{sec:naturalness}

A rejection-based rule lowers collision metrics partly by construction, so the reduction alone is weak evidence.
This subsection therefore asks three validity questions: are the selected names still natural, do the conclusions survive alternative counting conventions, and do they survive re-aggregation by language and model family.

\paragraph{Name naturalness.}
We matched every proper-name character against civil-registry lists of real given names: the US Social Security Administration names (1880--2008) for English, the Spanish INE register (names with frequency $\geq 20$) for Spanish, the French INSEE \emph{Fichier des prénoms} (1900--2020) for French, and the Berlin newborn register (2023) for German.
Matching uses the normalized first given name, the same convention as the tracker.
Table~\ref{tab:naturalness} reports the share of characters whose name is attested in the registry of its language.

\begin{table}[t]
\centering
\caption{Share of characters (\%, mean across 20 models) whose first given name is attested in a civil-registry list for the story language. List coverage differs across languages, so only within-language comparisons are meaningful. Static banlists push models toward invented names; speculative naming stays close to baseline.}
\label{tab:naturalness}
\small
\begin{tabular}{lcccc}
\toprule
Strategy & English & Spanish & French & German \\
\midrule
Baseline & 52.2 & 65.0 & 75.7 & 42.7 \\
Anti-slop prompt & 45.6 & 62.6 & 80.6 & 42.2 \\
Ban top 20 & 20.2 & 26.8 & 41.3 & 8.7 \\
Multiphase & 39.1 & 60.3 & 72.9 & 36.9 \\
Multiagent & 37.7 & 51.4 & 63.3 & 20.8 \\
Speculative $\lambda=0$ & 42.7 & 48.5 & 66.2 & 38.7 \\
Speculative $\lambda=10$ & 44.0 & 53.9 & 67.0 & 37.0 \\
Speculative HARD & 43.7 & 53.8 & 68.6 & 41.9 \\
\bottomrule
\end{tabular}
\end{table}

The banlist more than halves the share of real names in English and Spanish and collapses it in German (42.7\% to 8.7\%): when familiar names are forbidden, models reach for invented ones.
Speculative HARD stays within 0.8 to 11.2 percentage points of baseline, with the German difference smallest, and its most frequent non-attested names are fantasy-register coinages (\emph{Elowen}, \emph{Eira}, \emph{Caspian}) and the emergent attractor \emph{Kaelen}.
Attestation is not appropriateness: whether the names also fit the character and setting is a judgment question that we address with a blinded human evaluation (Section~\ref{sec:human-eval}).

\paragraph{Counting conventions.}
Analyses count full normalized names (\emph{Kaelen Voss} and \emph{Kaelen Thorne} are different names); the tracker scores first given names, so the two share one penalty.
Under the stricter first-token convention the speculative advantage narrows (HARD 0.0100 versus multiagent 0.0108, as 39 to 43\% of speculative names are compound, against 3 to 5\% at baseline) but HARD remains first or tied-first under either convention.

\begin{figure}[t]
  \centering
  \includegraphics[width=\textwidth]{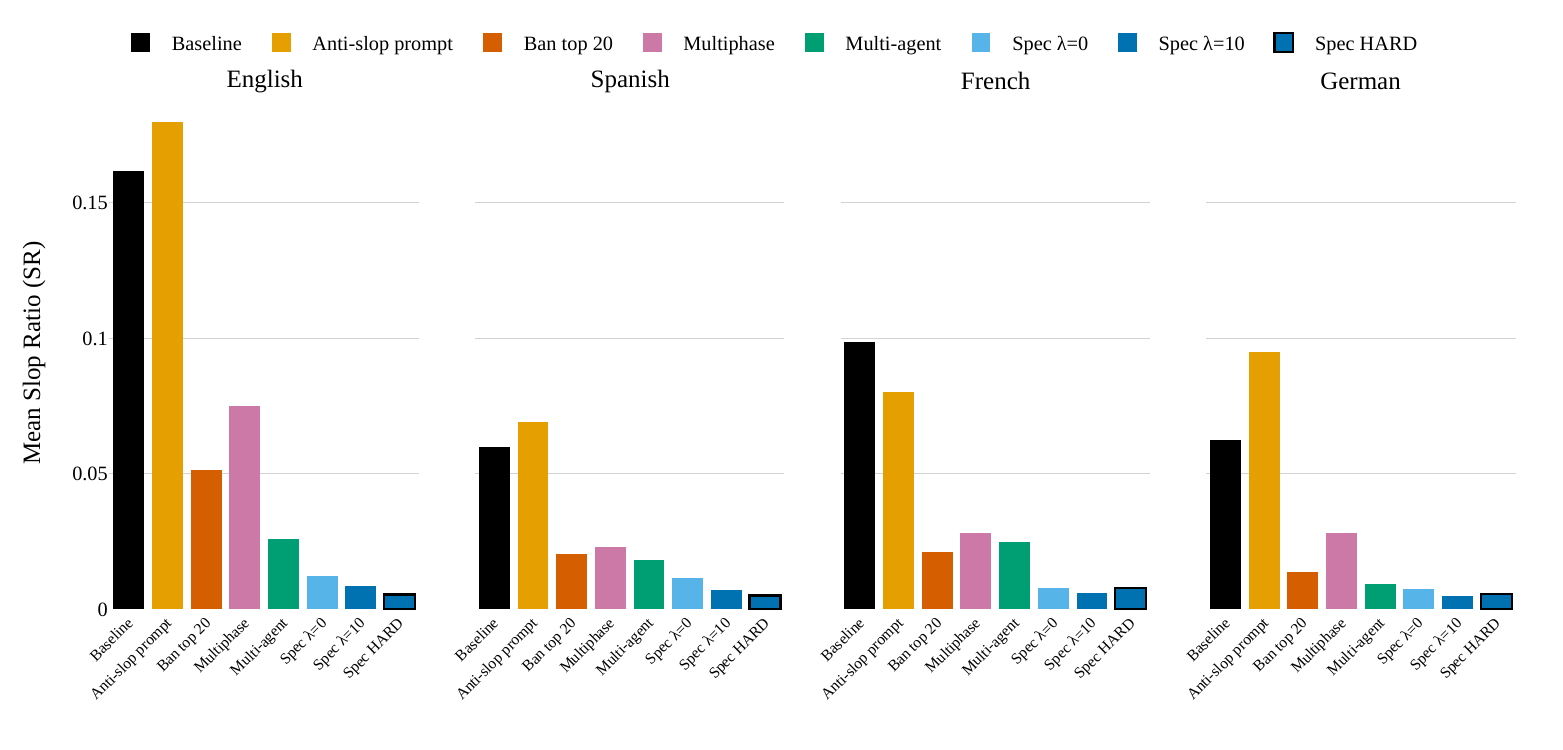}
  \caption{Mean Slop Ratio by story language for the eight strategies (20 models; Llama 3.3~70B excluded). English concentrates names more than the other languages at baseline, but the strategy ordering is stable across languages: speculative HARD (highlighted) has the lowest SR in all four.}
  \label{fig:language_efficacy}
\end{figure}

\paragraph{Re-aggregation.}
The ranking by mean SR is identical on English-only stories (Spearman $\rho = 1.000$) and when any model family is left out (minimum $\rho = 0.976$): speculative HARD ranks first in all ten views, with the lowest SR in each of the four languages (Figure~\ref{fig:language_efficacy}).

\subsection{Blinded human evaluation}\label{sec:human-eval}
Because the Slop Ratio is also the quantity speculative naming optimizes, we checked its conclusions with readers.
Three annotators evaluated the 44 pairs blind to strategy: the first author and two students unaffiliated with the project, one with a technical background and one without.
The set comprises the 36 main pairs of the three contrasts (2 per cell of model and language, with Llama 3.3 as a no-slop control) plus 8 diagnostic pairs and 2 attention checks, with three forced choices per pair: which story is more creative, which they would rather read, and in which story the character names feel more natural.
Judgments were merged by majority vote (2 of 3; a split counts as tie), and agreement was good on all three questions (Krippendorff $\alpha = 0.80$, 95\% CI $[0.67, 0.91]$ for names, $0.78$ for creativity, and $0.75$ for reading preference).

On the names question, consensus preferred speculative HARD in 6 of 8 decided comparisons against baseline (4 ties, exact binomial $p = 0.29$), preferred or tied it in 18 of the 24 pairs in which it appears, and preferred it in 4 of 8 decided comparisons against the multiagent reviewer (4 ties).
No question in any contrast shows a significant preference against speculative naming; the strongest baseline-leaning result, creativity against speculative HARD (7 of 10 decided for baseline), has $p = 0.34$. The only significant preference in the set favors another mitigation: the multiagent reviewer over baseline on reading preference (8 of 9 decided, $p = 0.039$).
Llama 3.3, whose baseline output rarely names characters, produced majority ties on the names question in all 4 of its baseline comparisons, as expected.
The four no-name-versus-named diagnostic pairs support the exclusion rule the analyses rely on (consensus never preferred the unnamed story on names), whereas the four stimulus-name pairs show that readers do not penalize stimulus-word names the way the extractor flags them.
The extractor's no-name flag, on which the analyses' exclusions rest, was blind-confirmed on all 19 flagged stories by the author annotator and on 15 by an independent reader; both readers confirmed only 1 of the 3 stimulus-flagged stories.

\subsection{Second-layer slop varies in depth (RQ4)}

Banning the top 20 names does not free the distribution; it shifts concentration to new names.
Table~\ref{tab:second_layer} (Appendix~\ref{app:A}) quantifies this second-layer slop across strategies, averaged over the 19 comparable models (per-model top-5 examples are in the supplementary material).
Three patterns stand out.
First, baseline and anti-slop leave the top-1 name at 21 to 23\% of all characters, and that name belongs to the canonical slop set in 89 to 95\% of models.
Second, ban top 20 cuts the top-1 share to 8\% and the slop rate to 26\%, but one in four models still has a slop name as its most frequent character: \emph{Hemlock} in Gemma, \emph{Fitzwilliam} in the antislop variant, \emph{Rowan} in GPT-5.4 Nano.

Multiphase is worse: \emph{Elara} leaks through its two-pass scaffold and remains top-1 in 8 of 19 models (42\% slop rate). Speculative HARD achieves the lowest top-1 share (2.6\%), the lowest $C_5$ (0.095), and a slop rate of only 5\% (1 of 19 models, \emph{Elena} for GPT-5.4 Nano).

The specific attractor names, however, vary enormously by model.
Tables~\ref{tab:A1}--\ref{tab:A3} (Appendix~\ref{app:A}) give the name-level evidence for every model: Gemma~3~27B shifts from deep slop (\emph{Tiber}) to second-layer slop (\emph{Hemlock}), Ministral~3~14B from \emph{Elias} to \emph{Veyra}, and Qwen~3.6~35B from \emph{Elara} to \emph{Zog}.
In each case, ban top 20 replaces one concentrated attractor with another, while speculative HARD produces genuinely new, low-concentration names.
One model inverts the expected effect: Llama~3.3~70B, which emits proper names in only 12\% of stories at baseline, starts naming aggressively when given a banlist and collapses onto two fantasy names (Zorvath 78 times, Zephyrine 73 times), raising its SR from 0.028 to 0.081: the instruction to avoid names implicitly signals that naming matters.

\subsection{The anti-slop prompt: a weak, non-significant rebound}

The generic instruction to avoid clich\'ed names does not improve any diversity metric: it slightly increases SR ($+0.011$, 95\% CI $[-0.005, +0.026]$, $p_{\text{Holm}} = 0.39$), worsens SR in 13 of 20 models, and wins zero of 100 comparisons.
Naming what to avoid may raise its probability, and the effect is too weak to help at the aggregate level.
By contrast, ban top 20, a concrete forbidden list, significantly reduces SR ($-0.051$, $p_{\text{Holm}} < 0.002$): the format of the prohibition matters.

\subsection{Reasoning mode does not reduce proper-name slop (RQ6)}
\label{sec:reasoning}

Four model families have both a base and a thinking variant, which enables a controlled comparison: thinking mode does not reduce proper-name slop, and if anything increases it ($+0.1$ to $+2.6$ percentage points across the four pairs).
An apparent benefit in earlier measurements came from role-based names and disappears on proper names only.
Speculative HARD remains the best strategy regardless of reasoning mode.
The full per-family breakdown (SR, $C_5$, entropy, unique names) is in Table~\ref{tab:B1} (Appendix~\ref{app:B}).

\subsection{Text versus vision modality (RQ2)}
\label{sec:modality}

Five model families have both text and vision variants.
Vision mode reduces name slop in four of five families, dramatically so for the Qwen models (42 to 90\% SR reduction across the three Qwen variants).
The visual stimulus appears to ground the narrative in concrete imagery, which diversifies character naming.
Gemma~3~27B is the exception.
Its slop is so deeply embedded that the input modality has a negligible effect.
Speculative HARD remains the best strategy in both modalities.
The full per-family breakdown is in Table~\ref{tab:C1} (Appendix~\ref{app:C}).

\subsection{Slop dialects by language (RQ5)}
\label{sec:dialects}

Slop names are partially language-specific.
Table~\ref{tab:languages} (Appendix~\ref{app:E}) sets the most frequent generated names per language, at baseline and under speculative HARD, against the five most popular names in the corresponding civil registry.
\emph{Elara}, \emph{Elias}, and \emph{Leo} appear in the generated top-5 of all four languages, but \emph{Elara} is the only one in the top-2 everywhere; \emph{Tiber} is English-only (and only for Gemma).
The comparison with the registries shows that the concentration is not ordinary popularity: the generated and registry top-5 are disjoint in every language except French, where \emph{Leo} leads both (at 16.6\% of generated characters versus 0.7\% of births).
The names that count as ``attested'' in Section~\ref{sec:naturalness} are therefore mostly low-frequency registry names, not the names real parents give at scale. The overlap with the registry's top-100 (bold in the table) is confined to mid-tier names at 0.3--1.8\% of registry volume, and no generated name in English reaches it.

The per-model breakdown (Table~\ref{tab:E1}, Appendix~\ref{app:E}) shows how far each fingerprint travels: Tiber leads Gemma's English output (67 versus 15 occurrences) and surfaces in German as \emph{Tiberius}, but not in Spanish or French; Gemini Flash Lite's \emph{Elias}, by contrast, leads in all four languages. French is dominated by \emph{Leo} and \emph{L\'ea}, German by \emph{Elias}, Spanish by \emph{Leo} and \emph{Luna}.
The cross-language Jaccard overlap of the top 50 names is consistently low (mean 0.094 at baseline).
Figure~\ref{fig:cross_language} (Appendix~\ref{app:E}) shows that speculative HARD and multiagent reduce this overlap further (to 0.056 and 0.046 respectively), meaning their output name sets diverge even more across languages.
Speculative HARD and multiagent reduce slop in all four languages, the anti-slop prompt worsens it in all four: monolingual banlists do not transfer, which supports the list-free design of SR.

\subsection{Name diversification is cost-free}

Lexical diversity is essentially invariant across strategies (Distinct-2 within 0.008 of baseline; Table~\ref{tab:lexical}, Appendix~\ref{app:A}): name-slop mitigation does not trade off against prose-level diversity.
Inter-story semantic similarity moves with the names themselves: speculative $\lambda=10$ reduces it in 14 of 20 models (mean $-0.013$), with the largest drops in the deepest-slop models, exactly where names are substituted most.
The measure does not establish plot diversification.
The deeper convergence question is examined in a companion study, which shows that without stimulus pictograms stories converge regardless of naming (mean cosine 0.312 versus 0.271; near-duplicate rate 38\% to 71\%).

\subsection{Cost-effectiveness (RQ3)}

Cost is part of the evaluation, not an afterthought, because any mitigation that doubles or triples the token budget may be unusable in production.
Table~\ref{tab:cost_effectiveness} (appendix) reports the cost of each strategy together with its Slop-Ratio reduction per 1{,}000 tokens and per model call.

\begin{figure}[t]
  \centering
  \includegraphics[width=0.85\textwidth]{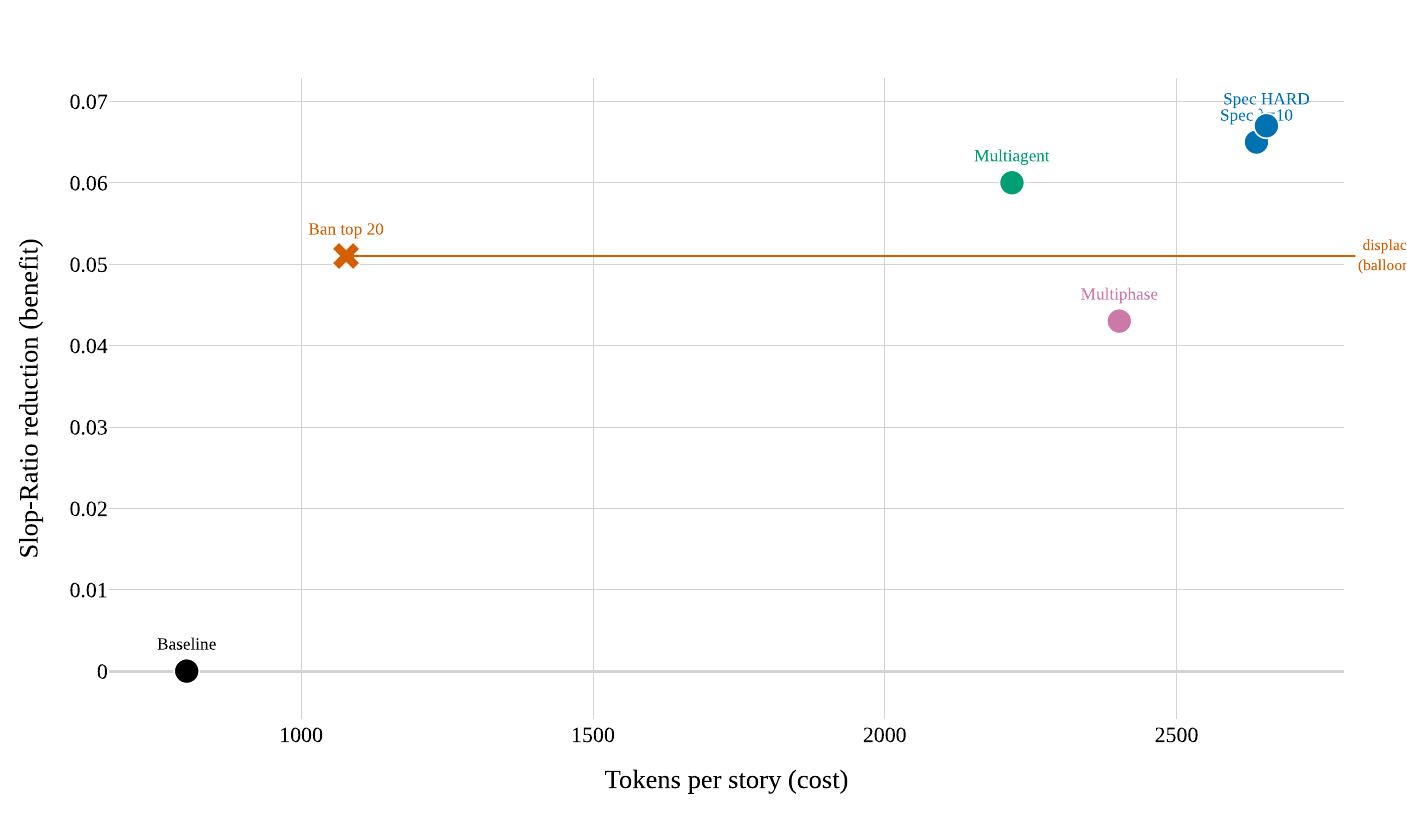}
  \caption{Slop-Ratio reduction against tokens per story, one point per strategy; up and to the left is better. Speculative HARD and $\lambda=10$ (blue) give the largest reduction at a fixed cost of about 2{,}650 tokens; ban top 20 (red cross) is cheap but only displaces slop.}
  \label{fig:cost_effectiveness}
\end{figure}

Ban top 20 has the best raw reduction-per-token ratio, but the emergent-slop column shows that its cheapness buys displacement rather than removal. Speculative HARD has the largest absolute and per-call reduction at a fixed, budgetable 3.30 times the baseline cost, unlike the variable-cost multiagent (470 to 16,885 tokens per story). Figure~\ref{fig:cost_effectiveness} plots the trade-off.

\subsection{Statistical significance of all strategies}

\begin{figure}[t]
  \centering
  \includegraphics[width=\textwidth]{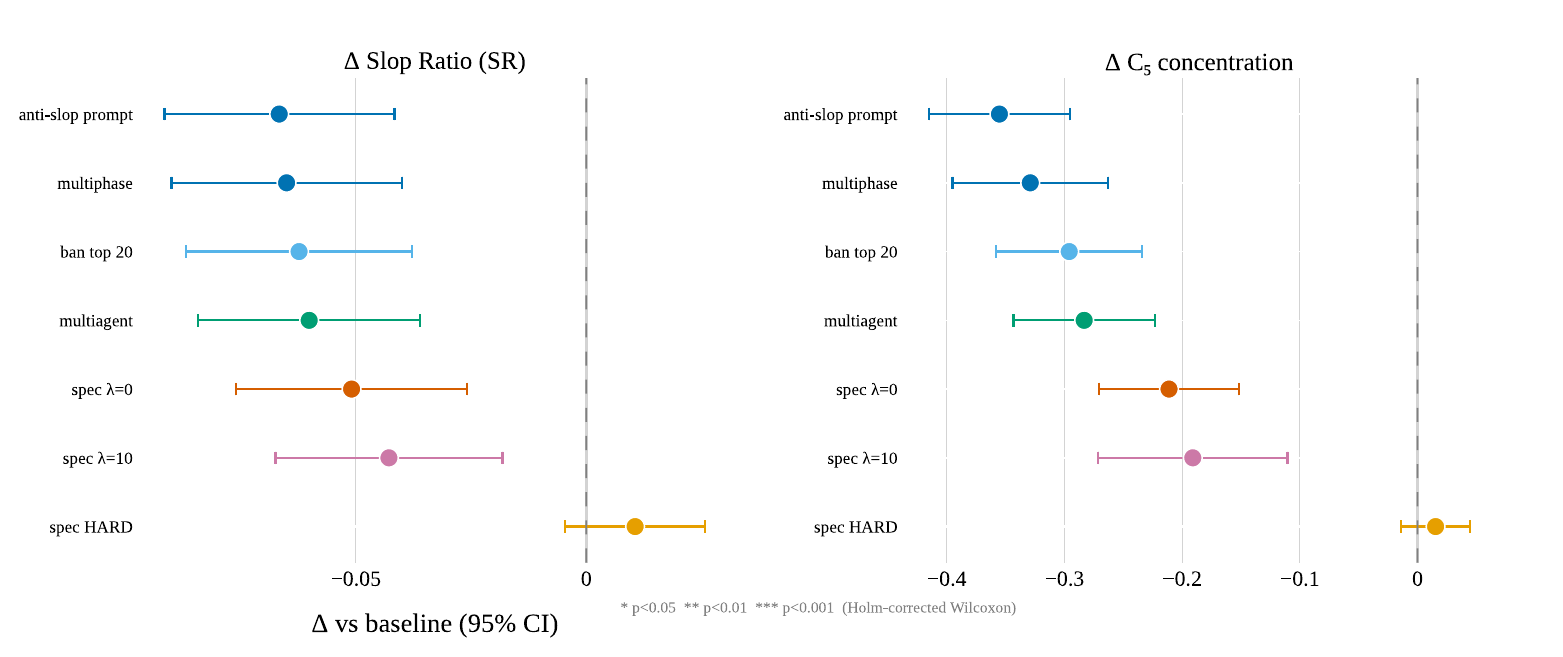}
  \caption{Forest plot of effect sizes: $\Delta$SR (left) and $\Delta C_5$ (right), each point the mean change across 20 models with 95\% CIs. Significance markers: * $p < 0.05$, ** $p < 0.01$, *** $p < 0.001$ (Holm-corrected). The anti-slop prompt (orange) is the only strategy whose interval crosses zero.}
  \label{fig:forest_plot}
\end{figure}

Figure~\ref{fig:forest_plot} summarizes the statistical evidence for every strategy on the two key slop metrics (SR and $C_5$).
All active strategies except the anti-slop prompt produce large, highly significant reductions (Holm-corrected Wilcoxon $p < 0.005$, Cohen's $|d|$ from 0.8 to 1.3 for SR and from 1.1 to 2.8 for $C_5$).
Speculative HARD has the largest effect on SR ($d = -1.25$, $\Delta = -0.067$, 95\% CI $[-0.092, -0.042]$) and on $C_5$ ($d = -2.78$).
The anti-slop prompt is the sole non-significant strategy ($p_{\text{Holm}} = 0.39$ for SR, $d = 0.33$), confirming that a vague negation instruction does not reliably reduce name concentration.
Table~\ref{tab:stats} in the appendix reports the full statistics.

\subsection{Speculative dose-response}

\begin{figure}[t]
  \centering
  \includegraphics[width=0.85\textwidth]{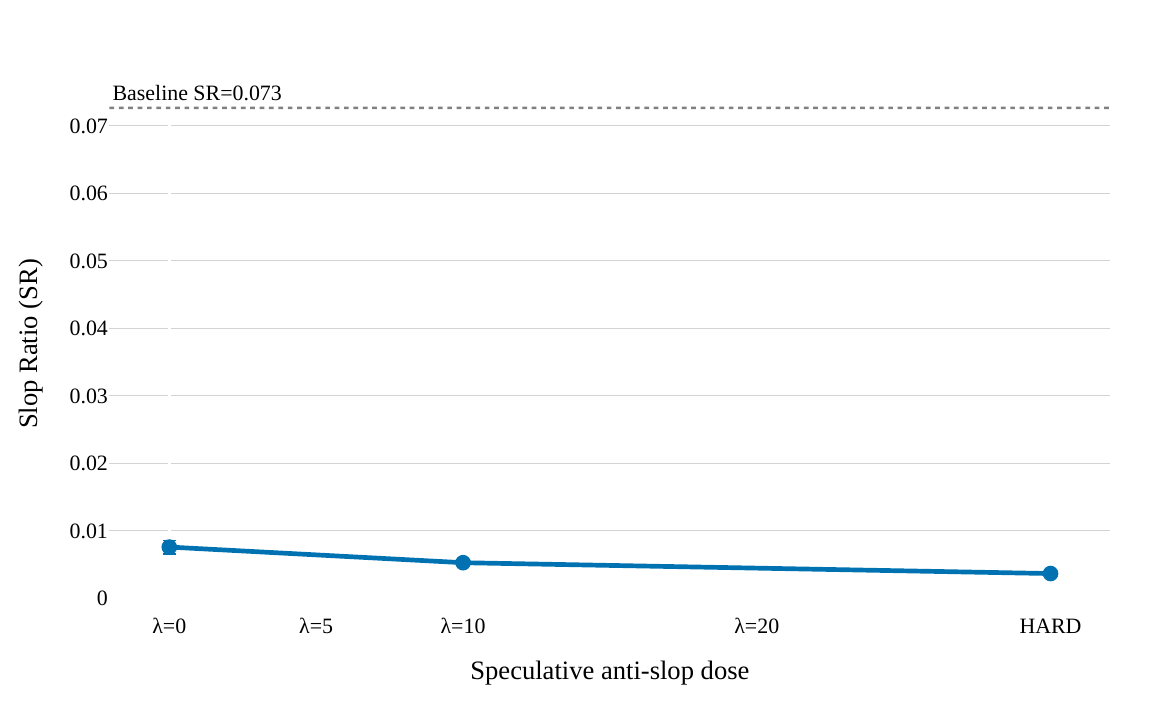}
  \caption{Mean Slop Ratio versus penalty $\lambda$ across 19 models (Llama 3.3 70B excluded); dotted line is baseline SR. Even $\lambda = 0$ reduces baseline SR by 89\%; the tracker adds monotonic suppression up to HARD. Error bars: one standard error.}
  \label{fig:dose_response}
\end{figure}

The speculative method has one parameter: the anti-slop penalty $\lambda$.
Figure~\ref{fig:dose_response} traces the Slop Ratio as $\lambda$ increases from 0 (no penalty, pure verbalized sampling) through 10 (moderate) to HARD (the strict threshold rule of Section~\ref{sec:method}).
Even at $\lambda = 0$, speculative naming reduces the baseline SR by 89\% (0.073 to 0.008), because the verbalized-sampling scaffold already diversifies candidate name-sets.
The tracker adds the last mile, and it matters most where slop is deepest: in deep-slop models, verbalized sampling alone leaves a 2 to 3 times higher SR than HARD (e.g.\ Qwen 3.5: 0.021 versus 0.007) and nearly double the top-1 share (4.6\% versus 2.6\%). It halves the emergent component and drives the known one to zero, as expected under a rejection rule.

\begin{figure}[t]
  \centering
  \includegraphics[width=\textwidth]{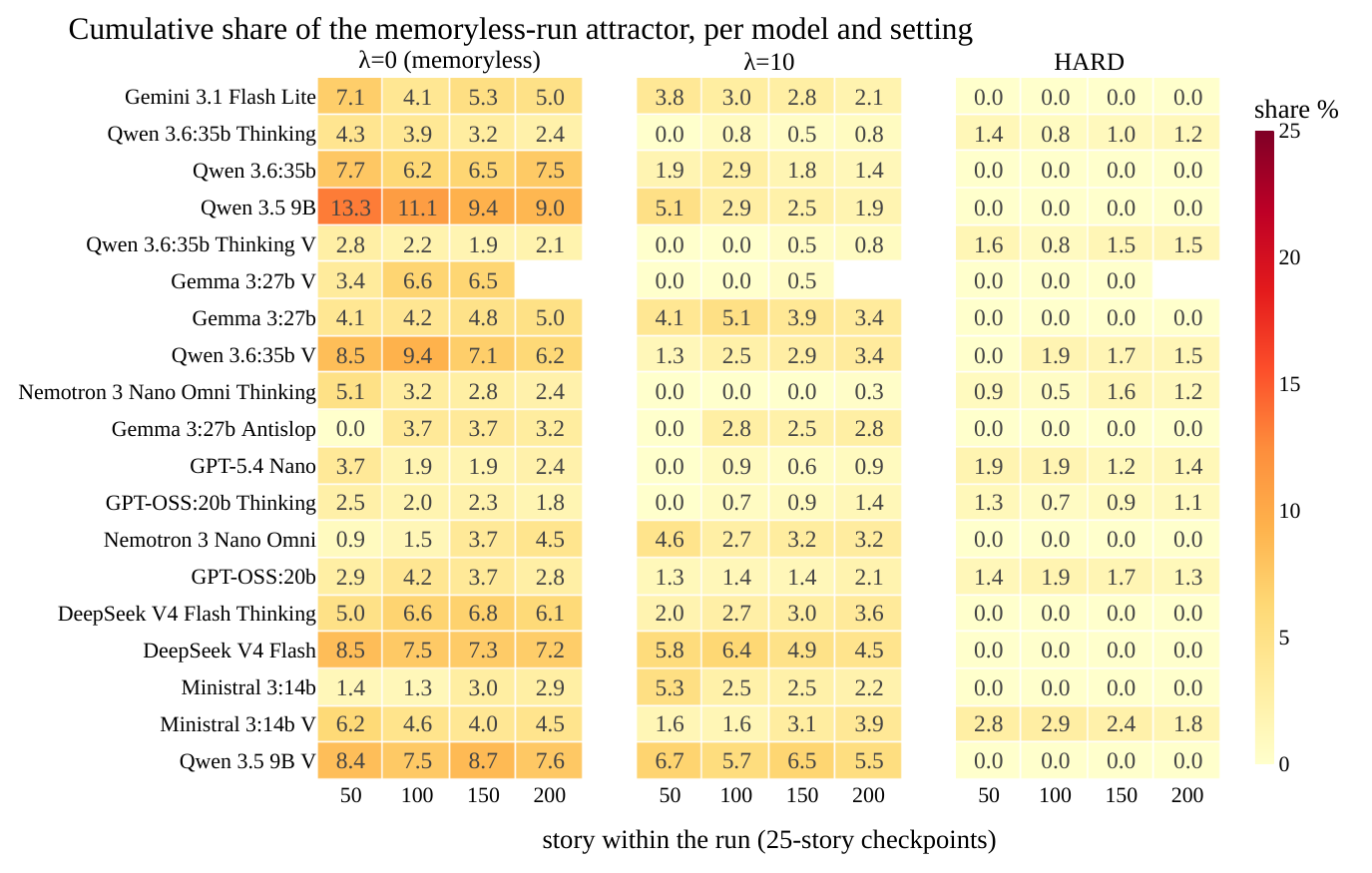}
  \caption{Eviction of the run-dominant attractor, per model and setting. For each model we track the name that dominates its memoryless ($\lambda=0$) run (\emph{Elara} for the Qwen and Gemma cells, \emph{Elias Thorne} for Gemini Flash Lite). Cells: cumulative share of that name, counting from story 1, at 50-story checkpoints (rows: the 19 models sorted by baseline SR; one decimal; $0.0$ = the name is never selected in the run, rejected at every choice; small nonzero values leak through the fallback). Under $\lambda=0$ the share persists for the whole run; under the tracked settings it collapses within the first checkpoints. The final share under HARD ends below the $\lambda=0$ level in all 19 models.}
  \label{fig:name_trajectory}
\end{figure}

The memory term is also what separates the method from verbalized sampling as such: no value of $\lambda$ is equivalent to $\lambda=0$, because every $\lambda>0$ activates the runtime counter and turns memoryless sampling into a self-correcting process.
The per-story trajectories show the difference directly (Figure~\ref{fig:name_trajectory}).
Under $\lambda=0$, the Qwen 3.5 9B attractor \emph{Elara} stays the top name for the entire 209-story run, at a stable 9 to 12\% of characters: memoryless sampling reproduces the verbalized distribution's own bias indefinitely.
Under the tracked settings \emph{Elara} is evicted within the same run, leaving a lower, rotating residual attractor (the emergent \emph{Kaelen} at 4 to 6\%). The pattern is systematic: tracking each model's memoryless-run attractor, the final share under HARD ends below the $\lambda=0$ level in all 19 models.

The SlopTracker adds incremental suppression: mean SR drops from 0.008 ($\lambda=0$) to 0.005 ($\lambda=10$) to 0.004 (HARD), a monotonic improvement.
The 7-model subset with intermediate $\lambda$ values (5 and 20) confirms the curve is smooth and monotonic (0.009, 0.007, 0.006, 0.006, 0.004) with no threshold or discontinuity. The per-model curves (Figure~\ref{fig:dose_per_model}, Appendix~\ref{app:A}) show the same monotone shape model by model, with the drop proportional to each model's baseline slop.
The practical recommendation is HARD when the goal is to remove attractors outright; Section~\ref{sec:practical} discusses when the intermediate $\lambda$ is the better operating point.

\section{Discussion}
\label{sec:discussion}

\subsection{The balloon effect}

The central conceptual contribution is the balloon effect, made precise by the SR decomposition.
A banlist compresses the name distribution at one point; the model shifts to the next available attractor, which then concentrates.
Because slop is a property of the aligned distribution rather than of a finite token set, every banlist is incomplete; the same pattern reaches fine-tuning, where the FTPO model of \citet{Paech2025antislop} suppresses its banlist names while new ones grow.
Dynamic tracking penalizes whatever begins to concentrate, whether or not it appeared in baseline.

Our own corpus contains a controlled demonstration: under $\lambda=10$ the emergent attractor \emph{Kaelen} still settles as top name in several models, while HARD rejects it and forces rotation of its surnames (\emph{Kaelen Voss}, \emph{Kaelen Thorne}, \emph{Kaelen Vane}; Table~\ref{tab:A3}, Appendix~\ref{app:A}). The rebound occurs even with a tracker present; what prevents it is the strength of the rejection, not merely the presence of state.

Two qualifications keep this reading honest.
First, most of the reduction is scaffold credit: verbalized sampling with placeholders already removes 89\% of baseline slop at $\lambda=0$; the tracker adds the removal of the known component (expected under a rejection rule) and a further halving of the emergent one.
Second, a rejection-based rule wins collision metrics partly by construction, which is why the naturalness and robustness checks of Section~\ref{sec:naturalness} carry evidential weight: the reduction does not come at the cost of real-name usage, and the ranking survives re-aggregation and counting conventions.

\subsection{Practical recommendations}
\label{sec:practical}

Speculative HARD is the default when a baseline fingerprint exists: it gives the lowest SR at a fixed cost of about three times the baseline token budget, on deep-slop local models and on API models alike (87 to 98\% SR reduction on Gemini Flash Lite, GPT-5.4 Nano, and DeepSeek Flash).
Multiagent is the fallback without a fingerprint, because its threshold needs no prior knowledge.
Within the speculative settings, $\lambda$ is an operating point, not a correctness question: HARD caps reuse at roughly one occurrence per run, below the natural reuse of human fiction, whereas $\lambda=10$ keeps a controlled residual.
We recommend HARD to remove attractors and $\lambda=10$ to approach natural concentration levels.

\subsection{Name slop as one case: toward other slop units}
\label{sec:beyond-names}

Name slop is the unit this paper measures and mitigates, but it is one instance of a broader problem: the same aligned models repeat sensory clich\'es (``a shimmer of light''), intensifiers (``profoundly''), formulaic openings (``In a world where''), and metaphors (``tapestry of emotions'') at rates far above human text~\citep{Paech2025antislop, Reinhart2025llms}.
The SlopTracker and the Slop Ratio are defined over any countable element, so the framework transfers to these units with one change: the extraction step.
We state this transfer as a hypothesis, not a result.
If a unit can be extracted from finished text (an opening line, a sensory phrase, a formulaic construction), the candidate-request and penalized-selection steps of Section~\ref{sec:method} apply unchanged.
Two conditions make the transfer non-trivial.
First, rewriting a phrase restructures the sentence, unlike a name substitution, so preserving prose quality becomes an empirical question for each new unit.
Second, the balloon-effect prediction (banning ``shimmer'' should grow ``glimmer'', just as banning \emph{Elara} grew \emph{Hemlock}) would need to be confirmed per unit.
We leave both to future work.

We measured this transfer for one additional unit, sentence openers.
Sentence-initial four-grams concentrate as names do: the dominant story opener accounts for 2.7--4.4\% of openings at the median model (22\% at the maximum).
Mid-story pivots recur across independent models: the dominant Spanish pivot appears in 20 of 20 models, the French one in 19 of 20.
Human review of the 80 top candidate openers in English and Spanish confirmed 94\% as slop; the legitimate remainder are genre conventions such as ``once upon a time''.
Name-targeted and general anti-clich\'e instructions leave this concentration largely intact or worsen it: the anti-slop prompt raises one cell from 32.5\% to 53.8\%.
Only the Gemma Antislop fine-tune reduces it, halving English opener concentration while the Spanish attractor survives unchanged.
The candidate-request and penalized-selection steps would apply unchanged to this unit; only the extraction step changes (Table~\ref{tab:opener_poc}).

\subsection{Limitations}
\label{sec:limitations}

This work has several limitations that stay with the study.
First, we demonstrate the method on one slop unit.
Section~\ref{sec:beyond-names} states the hypothesis that carries it to other units, but we do not run that transfer here.

Second, our evaluation is automated.
Each cell is also a single generation run of about 200 stories; the bootstrap intervals capture sampling variance within a cell, not run-to-run variance across repeated generations.
\citet{Chakrabarty2025polish} show that language models perform near chance when evaluating writing, and \citet{Shaib2025measuring} confirm that they fail to detect slop.
The blinded human evaluation of Section~\ref{sec:human-eval} partially addresses this: it includes two readers unaffiliated with the project and shows substantial agreement on the names question, but its scale (44 pairs, three readers) remains limited, and a larger study with paid external participants remains open.

Third, we evaluate a fixed element-order constraint, which controls structure across conditions but reduces ecological validity relative to free-order generation.
Fourth, strategy rankings vary by model family, so the recommendations are conditioned on the 20 models studied.

Fifth, the SlopTracker has a cold-start window.
The runtime counter activates only after 50 accumulated names, so early stories receive no emergent-slop penalty.
Measured directly, early-window SR exceeds late-window SR in 17 of 20 models (mean $\Delta$SR $= +0.0094$, 95\% CI $[+0.0038, +0.0160]$, $p < 0.001$), entirely emergent; the effect is bounded (early-window 0.012, below every other strategy except multiagent, decaying to 0.003 after warm-up).

A second tracker limitation concerns within-set collisions.
Candidate names within one story are scored independently, so two placeholders can receive names with the same normalized first token (\emph{Kaelen Voss} and \emph{Kaelen Thorne}).
The collision enters the runtime counter after selection and is penalized in later stories, but the story where it occurs keeps both names.
The case is rare (0.17\% of HARD stories, 0.26\% of $\lambda=10$, versus 4.3\% of baseline, among stories with at least two named characters).
Sixth, we did not evaluate speculative naming without stimulus pictograms.
In that condition, run for the 15 chat models only, stories converge regardless of naming: mean inter-story similarity reaches 0.312 without stimulus versus 0.271 with it, and the anti-slop prompt leaves it at 0.309.
Free-generation homogenization extends beyond names and stays open.
Seventh, the design target is anti-concentration, not the matching of human name statistics.
HARD suppresses reuse below the rates at which human fiction recycles common names, so its output is more diverse than any natural corpus; the appropriate target concentration is an empirical question we leave open.
Eighth, the tracker controls name choice but not discourse-level reference: characters with two-part names are referred to by the full name at nearly every mention, where human fiction alternates full name, given name, and pronoun. Whether, in compound names, the repeated token is the component the model would have reused as slop requires a post-hoc discourse analysis that we leave open.
Ninth, compound names expose a whole-name scoring gap: the tracker scores complete strings, so a model can reuse a known attractor by adding a surname.
The reuse exists at baseline, where 15\% of compound names start with a fingerprint name (\emph{Luna Voss}, \emph{Elias Thorne}), and drops to 0.7\% under HARD (16 of 2,213 compounds, about half genuine reuses such as \emph{Elara Vance}, the rest accent-collision artifacts such as Mar\'ia/Mara).
More effective compound-name management, scoring name components rather than whole strings, would close this gap and is already part of the future-work agenda (Section~\ref{sec:future}).
The studies outlined in Section~\ref{sec:future} remain future work.

\subsection{Reproducibility and use of language models}
\label{sec:repro}

The generation code (prompt templates, speculative naming, multiphase, and multiagent pipelines, with the stimulus pictograms and per-model baseline fingerprints) is released at \texttt{github.com/amozo-es/character-name-concentration}, and the dataset (two CSV files with the 44{,}782 stories and the 88{,}919 character extractions) at \texttt{huggingface.co/datasets/amozo/character-name-concentration}.
All per-cell metrics, win counts, rankings, and statistical tests are computed from these two files by the evaluation code in the same repository.
Hyperparameters: $K = 5$ candidate name-sets per story; $\lambda \in \{0, 10, \text{HARD}\}$; multiagent threshold of three uses per name per model; banlist of 20 names.
Stories were generated between 2026-05 and 2026-07 via the OpenRouter and Ollama endpoints of Section~\ref{sec:models}; the online models have been updated since, so the released dataset is the definitive artifact.
A large language model (Claude, Anthropic) assisted with editorial tasks during drafting; all experimental design, analysis, claims, and verification of cited numbers were performed by the authors, and no LLM was used for ideation or for generating the experimental results.

\subsection{Future work}
\label{sec:future}

Four directions follow from this work.
First, validate the transfer hypothesis of Section~\ref{sec:beyond-names} on a second slop unit (formulaic openings, sensory phrases, or part-of-speech templates), stacking verifiers for several units on the same two-pass scaffold, whose shared drafting call should make the cost subadditive.
The balloon-effect prediction is concrete: banning ``shimmer'' should create a new attractor (``glimmer''), raising the emergent component under a banlist while dynamic tracking suppresses it.
Sentence-final formulas are the invasive limit, where replacement restructures the sentence, so prose-quality checks must accompany the demonstration.

Second, calibrate the control target.
The SlopTracker can penalize any name above a target rate $\tau$, calibrated to human-level name reuse, instead of penalizing any repetition at all; the zero-tolerance limit of that family (a HARD variant with $\tau = 0$) suppresses reuse below natural rates (Section~\ref{sec:practical}).
A calibrated $\tau$ would preserve natural name recurrence while still removing attractors, and a lower activation threshold or a sample-size-weighted penalty would shrink the cold-start window measured in Section~\ref{sec:limitations}.

Third, extend speculative naming to no-stimulus generation, where stories converge regardless of naming, combine the SlopTracker with the multiagent reviewer, and validate the rankings under free-order generation.
Fourth, extend control to the discourse level: score the parts of a multi-part name individually (so a rejected given name does not force the surname into every mention), alternate reference forms the way human fiction does (full name at introduction, given name and pronoun after), and run a post-hoc study of whether, in generated compound names, the repeated token is the model's slop component.

\section{Conclusion}
\label{sec:conclusion}

We presented an empirical study of name-concentration control in LLM fiction and introduced speculative naming, a fine-tuning-free method that adapts draft-verify-reject from the token level to the character-name level.
Across 20 models, four languages, and 44{,}782 stories, the sampling scaffold alone removes 89\% of baseline concentration, and the dynamic tracker brings the reduction to 95\% while holding known-name collisions at zero and keeping names close to registry naturalness.
Static banlists merely displace concentration, the balloon effect that the SR decomposition makes visible.
The scope of these results is name concentration; they do not establish gains in narrative creativity, and in no-stimulus generation stories converge regardless of naming, so plot-level homogenization remains open.
Section~\ref{sec:future} details the future-work directions, from second slop units to a calibrated control target and the human evaluation.


\bibliographystyle{unsrtnat}
\bibliography{references}


\appendix

\begin{table}[t]
\centering
\caption{Cost and cost-effectiveness of each active strategy. Tokens are cumulative across every model call for one story, measured from generation metadata; multiagent calls are 2 + revision rounds (0 to 3, average 2.65). Reduction is the absolute SR drop averaged across 20 models; ``Per 1K tokens'' divides it by the token cost of one story. The final column flags whether the strategy keeps emergent slop near zero or merely displaces it.}
\label{tab:cost_effectiveness}
\small
\begin{tabular}{lcccccc}
\toprule
Strategy & Calls & Tokens & $\times$ base & SR reduction & Per 1K tokens & Emergent slop \\
\midrule
Ban top 20 & 1 & 1{,}077 & 1.34 & 0.051 & 0.047 & Displaced \\
Multiagent & 2.65 avg & 2{,}218 & 2.76 & 0.060 & 0.027 & Reduced \\
Speculative $\lambda=10$ & 2 & 2{,}637 & 3.28 & 0.065 & 0.025 & Reduced \\
Speculative HARD & 2 & 2{,}654 & 3.30 & 0.067 & 0.025 & Near zero \\
Multiphase & 2 & 2{,}402 & 2.99 & 0.043 & 0.018 & Reduced \\
\bottomrule
\end{tabular}
\end{table}

\begin{table}[t]
\centering
\caption{Number of models (of 20) for which each strategy achieves the best value on each diversity metric. Arrows indicate the direction of improvement: lower is better for SR and $C_5$, higher is better for entropy, unique names, and evenness. The rightmost column sums each row to a maximum of 100. Once the speculative method is included, no standard strategy wins on either slop metric.}
\label{tab:wins}
\small
\begin{tabular}{lcccccr}
\toprule
Strategy & SR$\downarrow$ & $C_5\downarrow$ & $H\uparrow$ & $V\uparrow$ & $E\uparrow$ & Total \\
\midrule
\textbf{Speculative HARD} & 14 & 12 & 14 & 7 & 10 & \textbf{57} \\
Speculative $\lambda=10$ & 5 & 6 & 2 & 1 & 4 & \textbf{18} \\
Multiphase & 0 & 0 & 4 & 11 & 0 & \textbf{15} \\
Multiagent & 1 & 2 & 0 & 1 & 6 & \textbf{10} \\
Ban top 20 & 0 & 0 & 0 & 0 & 0 & \textbf{0} \\
Speculative $\lambda=0$ & 0 & 0 & 0 & 0 & 0 & \textbf{0} \\
Anti-slop & 0 & 0 & 0 & 0 & 0 & \textbf{0} \\
Baseline & 0 & 0 & 0 & 0 & 0 & \textbf{0} \\
\bottomrule
\end{tabular}
\end{table}

\begin{table}[t]
\centering
\caption{Second-layer slop by strategy, averaged across 19 models (Llama 3.3 70B excluded). ``Top-1 share'' is the mean percentage of characters assigned the single most frequent name. $C_5$ is top-5 concentration. $\mathrm{SR}_\text{emergent}$ is the emergent component of the Slop Ratio. ``Top-1 is slop'' is the percentage of models where the most frequent name belongs to the canonical slop set. The dagger ($\dagger$) marks strategies where slop-class names dominate the top-1 position.}
\label{tab:second_layer}
\small
\begin{tabular}{lccccc}
\toprule
Strategy & Top-1 share (\%) & $C_5$ & $\mathrm{SR}_\text{emergent}$ & Unique $V$ & Top-1 is slop (\%) \\
\midrule
Baseline & 20.9 & 0.455 & 0.002 & 75 & 89 $\dagger$ \\
Anti-slop prompt & 22.8 & 0.477 & 0.005 & 78 & 95 $\dagger$ \\
Ban top 20 & 7.9 & 0.227 & 0.014 & 154 & 26 \\
Multiphase & 10.2 & 0.255 & 0.011 & 209 & 42 \\
Multiagent & 5.0 & 0.160 & 0.008 & 129 & 21 \\
Speculative $\lambda=10$ & 3.4 & 0.120 & 0.004 & 195 & 32 \\
Speculative HARD & 2.6 & 0.095 & 0.004 & 206 & 5 \\
\bottomrule
\end{tabular}
\end{table}

\begin{table}[t]
\centering
\caption{Most frequent proper names per language, generated versus registered. For each language and rank: the generated names at baseline and under speculative HARD (share of that language's named characters, 19 models, Llama 3.3 70B excluded) and the five most popular names in the corresponding civil registry (SSA 2008 for English; INE cumulative counts for Spanish; INSEE 2020 births for French; Berlin 2023 for German). Shares are within-source and not comparable across columns. Bold marks generated names that fall within the registry's 100 most popular names of that language, with the registry share in parentheses. The generated top-5 and the registry top-5 are disjoint in every language except French, where \emph{Leo} is both the generated and the registry leader (at 16.6\% versus 0.7\%).}
\label{tab:languages}
\small
\begin{tabular}{llcccccc}
\toprule
& & \multicolumn{2}{c}{Generated (baseline)} & \multicolumn{2}{c}{Generated (HARD)} & \multicolumn{2}{c}{Registry} \\
\cmidrule(lr){3-4} \cmidrule(lr){5-6} \cmidrule(lr){7-8}
Language & \# & Name & \% & Name & \% & Name & \% \\
\midrule
\multirow{5}{*}{English} & 1 & Elara & 18.0 & Elowen & 1.9 & Jacob & 0.7 \\
 & 2 & Tiber & 12.3 & Kaelen & 1.2 & Michael & 0.6 \\
 & 3 & Leo & 7.5 & Soren & 0.9 & Ethan & 0.6 \\
 & 4 & Elias & 4.8 & Kaelen Thorne & 0.9 & Emma & 0.6 \\
 & 5 & Maya & 4.5 & Kael & 0.8 & Isabella & 0.6 \\
\midrule
\multirow{5}{*}{Spanish} & 1 & Leo & 9.2 & Mateo & 1.5 & Maria & 12.3 \\
 & 2 & Elara & 5.3 & Isolde & 1.1 & Jose & 5.5 \\
 & 3 & Elias & 5.0 & \textbf{Elena} (0.4) & 1.0 & Juan & 3.1 \\
 & 4 & Luna & 4.4 & Kael & 0.9 & Francisco & 2.5 \\
 & 5 & \textbf{Ana} (1.8) & 3.1 & Kaelen & 0.8 & Antonio & 2.1 \\
\midrule
\multirow{5}{*}{French} & 1 & \textbf{Leo} (0.7) & 16.6 & Elian & 1.7 & Leo & 0.7 \\
 & 2 & Elara & 7.1 & \textbf{Theo} (0.3) & 1.6 & Gabriel & 0.7 \\
 & 3 & \textbf{Lea} (0.4) & 6.6 & Elowen & 1.4 & Raphael & 0.6 \\
 & 4 & Elias & 6.4 & Lysandre & 1.3 & Louis & 0.6 \\
 & 5 & Lila & 2.6 & Eloi & 1.3 & Louise & 0.6 \\
\midrule
\multirow{5}{*}{German} & 1 & \textbf{Elias} (0.7) & 10.0 & \textbf{Lukas} (0.3) & 2.0 & Sophie & 1.0 \\
 & 2 & Elara & 8.4 & \textbf{Mira} (0.3) & 1.6 & Marie & 1.0 \\
 & 3 & \textbf{Lina} (0.5) & 5.7 & \textbf{Elias} (0.7) & 1.5 & Charlotte & 0.9 \\
 & 4 & \textbf{Lena} (0.3) & 5.3 & Mara & 1.3 & Maria & 0.9 \\
 & 5 & \textbf{Leo} (0.5) & 4.8 & \textbf{Lena} (0.3) & 1.2 & Emilia & 0.9 \\
\bottomrule
\end{tabular}
\end{table}

\begin{table}[t]
\centering
\caption{Lexical diversity by strategy, averaged across the 20 models. Distinct-2 is the type-token ratio over bigrams. PATTR is the positional-adjusted type-token ratio. MATTR-50 is the moving-average type-token ratio over a 50-token window. Higher is better for all three. All strategies fall within 0.008 of baseline on Distinct-2, which confirms that name-slop mitigation does not degrade lexical diversity.}
\label{tab:lexical}
\small
\begin{tabular}{lccc}
\toprule
Strategy & Distinct-2 & PATTR & MATTR-50 \\
\midrule
Baseline & 0.976 & 0.618 & 0.831 \\
Anti-slop prompt & 0.977 & 0.619 & 0.833 \\
Ban top 20 & 0.976 & 0.619 & 0.836 \\
Multiphase & 0.972 & 0.607 & 0.833 \\
Multiagent & 0.976 & 0.615 & 0.832 \\
Speculative $\lambda=0$ & 0.968 & 0.608 & 0.832 \\
Speculative $\lambda=10$ & 0.968 & 0.610 & 0.832 \\
Speculative HARD & 0.970 & 0.613 & 0.833 \\
\bottomrule
\end{tabular}
\end{table}

\begin{table}[t]
\centering
\caption{Statistical effects of each strategy versus baseline on the two slop metrics, across 20 models. $\Delta$ is the mean change (negative = improvement). CI is the bootstrap 95\% confidence interval. $d$ is Cohen's $d$. $p_{\text{Holm}}$ is the Holm--Bonferroni-corrected Wilcoxon signed-rank $p$-value. Significance flags: *** $p < 0.001$, ** $p < 0.01$, * $p < 0.05$, n.s.\ not significant.}
\label{tab:stats}
\footnotesize
\begin{tabular}{lccccccc}
\toprule
Strategy & $\Delta$SR & 95\% CI (SR) & $d$ (SR) & $p_{\text{Holm}}$ (SR) & $\Delta C_5$ & $d$ ($C_5$) & $p_{\text{Holm}}$ ($C_5$) \\
\midrule
Speculative HARD & $-0.067$ & $[-0.092, -0.042]$ & $-1.25$ & $<0.001$ *** & $-0.355$ & $-2.78$ & $<0.001$ *** \\
Speculative $\lambda=10$ & $-0.065$ & $[-0.090, -0.040]$ & $-1.21$ & $<0.001$ *** & $-0.329$ & $-2.33$ & $<0.001$ *** \\
Speculative $\lambda=0$ & $-0.062$ & $[-0.087, -0.038]$ & $-1.19$ & $<0.001$ *** & $-0.296$ & $-2.23$ & $<0.001$ *** \\
Multiagent & $-0.060$ & $[-0.084, -0.036]$ & $-1.17$ & $<0.001$ *** & $-0.283$ & $-2.20$ & $<0.001$ *** \\
Ban top 20 & $-0.051$ & $[-0.076, -0.026]$ & $-0.95$ & $0.001$ ** & $-0.211$ & $-1.67$ & $<0.001$ *** \\
Multiphase & $-0.043$ & $[-0.067, -0.018]$ & $-0.82$ & $0.003$ ** & $-0.191$ & $-1.11$ & $0.001$ ** \\
Anti-slop prompt & $+0.011$ & $[-0.005, +0.026]$ & $+0.33$ & $0.39$ n.s. & $+0.015$ & $+0.25$ & $0.46$ n.s. \\
\bottomrule
\end{tabular}
\end{table}

\begin{figure}[t]
  \centering
  \includegraphics[width=0.85\textwidth]{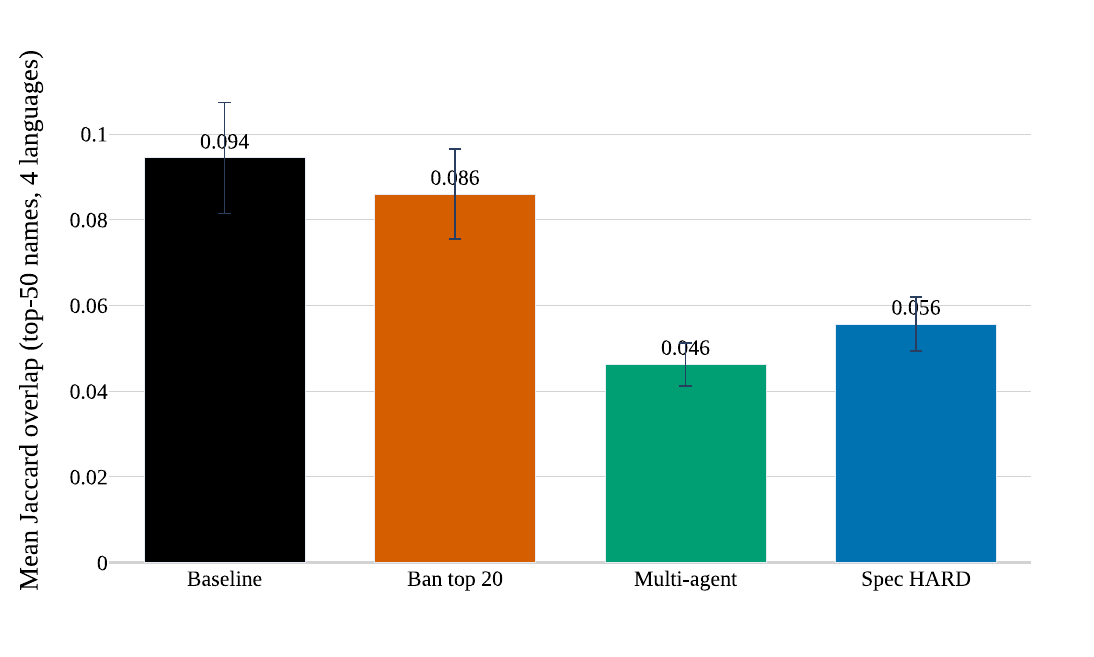}
  \caption{Mean Jaccard overlap of the top-50 character names across the six language pairs (en, es, fr, de), per strategy. Lower means more language-specific slop. Speculative HARD (0.056) and multiagent (0.046) produce the most language-divergent name sets, confirming that dynamic tracking generalizes across languages without a shared banlist.}
  \label{fig:cross_language}
\end{figure}

\section{Top-5 character names by model: baseline versus ban top 20}
\label{app:A}

Tables~\ref{tab:A1}--\ref{tab:A3} list the three most frequent proper names and their raw counts for every model under baseline, ban top 20 (the 20-name forbidden list), and speculative HARD.
Models are sorted by baseline SR (highest slop at top).
The balloon effect is visible in every row.
Ban top 20 suppresses the original attractors but replaces them with new ones that are often equally concentrated.
Gemma shifts from \emph{Tiber} to \emph{Hemlock} and \emph{Fitzwilliam}, which are themselves in the slop set.
Ministral shifts from \emph{Elias} to \emph{Veyra} (42 occurrences).
Qwen~3.6 shifts from \emph{Elara} to \emph{Zog} (30).
Gemini Flash Lite shifts from \emph{Elias} (75) to \emph{Zorion} and \emph{Thalric} (19 each).
Llama~3.3~70B, which barely names characters at baseline, produces extreme second-layer concentration under ban top 20: \emph{Zorvath} appears 78 times and \emph{Zephyrine} 73 times in approximately 200 stories, raising its SR from 0.028 to 0.081.

\begin{table}[t]
\centering
\caption{Top-3 character names per model under baseline, sorted by Slop Ratio (descending). SR is the Slop Ratio (lower is better). Counts are the number of characters assigned that name across approximately 200 stories.}
\label{tab:A1}
\scriptsize
\begin{tabular}{lccccc}
\toprule
Model & SR & \#1 & \#2 & \#3 \\
\midrule
Gemini 3.1 Flash Lite (text) & 0.188 & Elias (75) & Elara (26) & Elian (18) \\
Qwen 3.6:35b Thinking (text) & 0.177 & Elara (60) & Elias (17) & Leo (16) \\
Qwen 3.6:35b (text) & 0.151 & Elara (74) & Leo (24) & Barnaby (19) \\
Qwen 3.5 9B (text) & 0.135 & Elara (48) & Leo (20) & Barnaby (10) \\
Qwen 3.6:35b Thinking (vis) & 0.102 & Leo (44) & Elara (27) & Barnaby (9) \\
Gemma 3:27b (vis) & 0.092 & Tiber (70) & Leo (31) & Elara (20) \\
Gemma 3:27b (text) & 0.091 & Tiber (68) & Elara (42) & Elias (20) \\
Qwen 3.6:35b (vis) & 0.065 & Leo (35) & Elara (24) & Lila (11) \\
Nemotron 3 Nano Omni Thinking & 0.057 & Maya (37) & Lena (8) & Lina (8) \\
Gemma 3:27b Antislop (text) & 0.044 & Tiber (47) & Elias (18) & Flora (15) \\
GPT-5.4 Nano (text) & 0.038 & Mira (19) & Mara (9) & Lina (8) \\
GPT-OSS:20b Thinking (text) & 0.035 & Mara (16) & Maya (11) & Mira (10) \\
Nemotron 3 Nano Omni (text) & 0.035 & Elara (27) & Lila (11) & Leo (9) \\
GPT-OSS:20b (text) & 0.034 & L\'eo (13) & Mira (11) & Mara (8) \\
DeepSeek V4 Flash Thinking & 0.033 & Leo (21) & Elara (7) & Elias (5) \\
DeepSeek V4 Flash (text) & 0.032 & Leo (23) & Elara (18) & Elias (7) \\
Ministral 3:14b (text) & 0.032 & Elias (30) & Liora (16) & Lena (14) \\
Llama 3.3:70b (text) & 0.028 & Abenteurer (4) & Wizard (3) & Jack (2) \\
Ministral 3:14b (vis) & 0.026 & L\'eo (20) & Lina (18) & Lila (16) \\
Qwen 3.5 9B (vis) & 0.014 & Leo (8) & L\'eon (4) & Elara (4) \\
\bottomrule
\end{tabular}
\end{table}

\begin{table}[t]
\centering
\caption{Top-3 character names per model under ban top 20 (20-name forbidden list). Same model order as Table~\ref{tab:A1}. The original slop names are suppressed, but new attractors emerge with comparable or higher concentration: Hemlock (44) replaces Tiber in Gemma, Veyra (42) replaces Elias in Ministral, Zog (30) replaces Elara in Qwen 3.6. Llama 3.3 70B SR rises from 0.028 to 0.081 (Zorvath = 78).}
\label{tab:A2}
\scriptsize
\begin{tabular}{lccccc}
\toprule
Model & SR & \#1 & \#2 & \#3 \\
\midrule
Gemini 3.1 Flash Lite (text) & 0.022 & Zorion (19) & Thalric (19) & Z\'ephirin (12) \\
Qwen 3.6:35b Thinking (text) & 0.016 & Vael (16) & Vaelen (13) & Vaelis (10) \\
Qwen 3.6:35b (text) & 0.040 & Zog (30) & Zephyr (24) & Vex (17) \\
Qwen 3.5 9B (text) & 0.024 & Barnaby (25) & Vex (13) & Bramble (11) \\
Qwen 3.6:35b Thinking (vis) & 0.017 & Oryn (17) & Zog (17) & Bram (14) \\
Gemma 3:27b (vis) & 0.024 & Hemlock (39) & Fitzwilliam (31) & Wren (13) \\
Gemma 3:27b (text) & 0.030 & Hemlock (44) & Fitzwilliam (36) & Zephyr (24) \\
Qwen 3.6:35b (vis) & 0.010 & Zephyr (22) & Orin (12) & Zorin (10) \\
Nemotron 3 Nano Omni Thinking & 0.011 & Thalen (14) & Cassian (12) & Thalor (8) \\
Gemma 3:27b Antislop (text) & 0.025 & Fitzwilliam (51) & Bram (14) & Wren (13) \\
GPT-5.4 Nano (text) & 0.006 & Rowan (6) & Mara (4) & Maribel (3) \\
GPT-OSS:20b Thinking (text) & 0.006 & Z\'ephyr (7) & Nyx (7) & Mira (7) \\
Nemotron 3 Nano Omni (text) & 0.015 & Finn (15) & Kaelen (14) & Arin (10) \\
GPT-OSS:20b (text) & 0.007 & Vesper (9) & Zara (7) & Arion (7) \\
DeepSeek V4 Flash Thinking & 0.011 & Zephyr (22) & Finn (8) & Rook (5) \\
DeepSeek V4 Flash (text) & 0.014 & Zephyr (21) & Finn (16) & Z\'ephyrin (6) \\
Ministral 3:14b (text) & 0.017 & Veyra (42) & Dravik (11) & Orren (9) \\
Llama 3.3:70b (text) & 0.081 & Zorvath (78) & Zephyrine (73) & Xylara (34) \\
Ministral 3:14b (vis) & 0.007 & Orren (18) & Veyra (18) & Tansy (12) \\
Qwen 3.5 9B (vis) & 0.009 & Barnaby (16) & Zorin (10) & Bramble (7) \\
\bottomrule
\end{tabular}
\end{table}

\begin{table}[t]
\centering
\caption{Top-3 proper names per model under speculative HARD. Counts are raw character counts over the 200-story cell. Concentration collapses to an order of magnitude below baseline: top names hold 3 to 13 occurrences, versus 30 to 75 for the deep-slop models at baseline (Table~\ref{tab:A1}). Canonical slop names survive only at trivial counts and never as concentrated attractors (Elara for GPT-OSS:20b at 6; Elias only inside draft-style full names such as Elias Thorne for Gemma vision); the heads of the lists are draft-style combinations (Kaelen Voss, Elias Thorne) whose concentration lives in the shared first name, at 1/10th or less of the baseline scale.}
\label{tab:A3}
\scriptsize
\resizebox{\textwidth}{!}{%
\begin{tabular}{lccccc}
\toprule
Model & SR & \#1 & \#2 & \#3 \\
\midrule
Gemini 3.1 Flash Lite (text) & 0.188 & Kaelen Voss (8) & Kaelen Thorne (8) & Kaelen Vane (5) \\
Qwen 3.6:35b Thinking (text) & 0.177 & Elowen (5) & Maelis (4) & Unicorn (3) \\
Qwen 3.6:35b (text) & 0.151 & Elian Voss (11) & Elian Vane (7) & Kaelen (6) \\
Qwen 3.5 9B (text) & 0.135 & Kaelen (13) & \'Elian (10) & Theron (5) \\
Qwen 3.6:35b Thinking (vis) & 0.102 & Elowen (5) & Vaelen (5) & Maren Voss (4) \\
Gemma 3:27b (vis) & 0.092 & Elias Thorne (8) & Silas Blackwood (5) & Elowen (5) \\
Gemma 3:27b (text) & 0.090 & Alistair Finch (7) & Renzo Bellini (5) & Elowen (4) \\
Qwen 3.6:35b (vis) & 0.065 & Mira (7) & Nyx (6) & Thalor (6) \\
Nemotron 3 Nano Omni Thinking & 0.057 & Niamh (9) & Eira (8) & Mateo (8) \\
Gemma 3:27b Antislop (text) & 0.044 & Linnea (4) & Saoirse Byrne (4) & Lysandre (3) \\
GPT-5.4 Nano (text) & 0.038 & Elena (3) & Elara (3) & \'Elian (2) \\
GPT-OSS:20b Thinking (text) & 0.035 & Elowen (6) & Liora (3) & L\'eandre (3) \\
Nemotron 3 Nano Omni (text) & 0.035 & Axiom (7) & Kael (7) & Isolde (7) \\
GPT-OSS:20b (text) & 0.034 & Elara (6) & Lena (6) & Elowen (6) \\
DeepSeek V4 Flash Thinking & 0.033 & Kaelen (5) & Mira (5) & Caspian (4) \\
DeepSeek V4 Flash (text) & 0.032 & Kael (6) & Kai (5) & Finn (4) \\
Ministral 3:14b (text) & 0.032 & Eolande (7) & Eamon (6) & Kaelan (5) \\
Llama 3.3:70b (text) & 0.028 & Astrid Blackwood (8) & Kai Rylan (7) & Wren Everett (7) \\
Ministral 3:14b (vis) & 0.026 & Eolande (7) & Liora (5) & Orion (4) \\
Qwen 3.5 9B (vis) & 0.014 & Mira (8) & Kael (8) & Elian (8) \\
\bottomrule
\end{tabular}%
}
\end{table}

\begin{figure}[t]
\centering
\includegraphics[width=\textwidth]{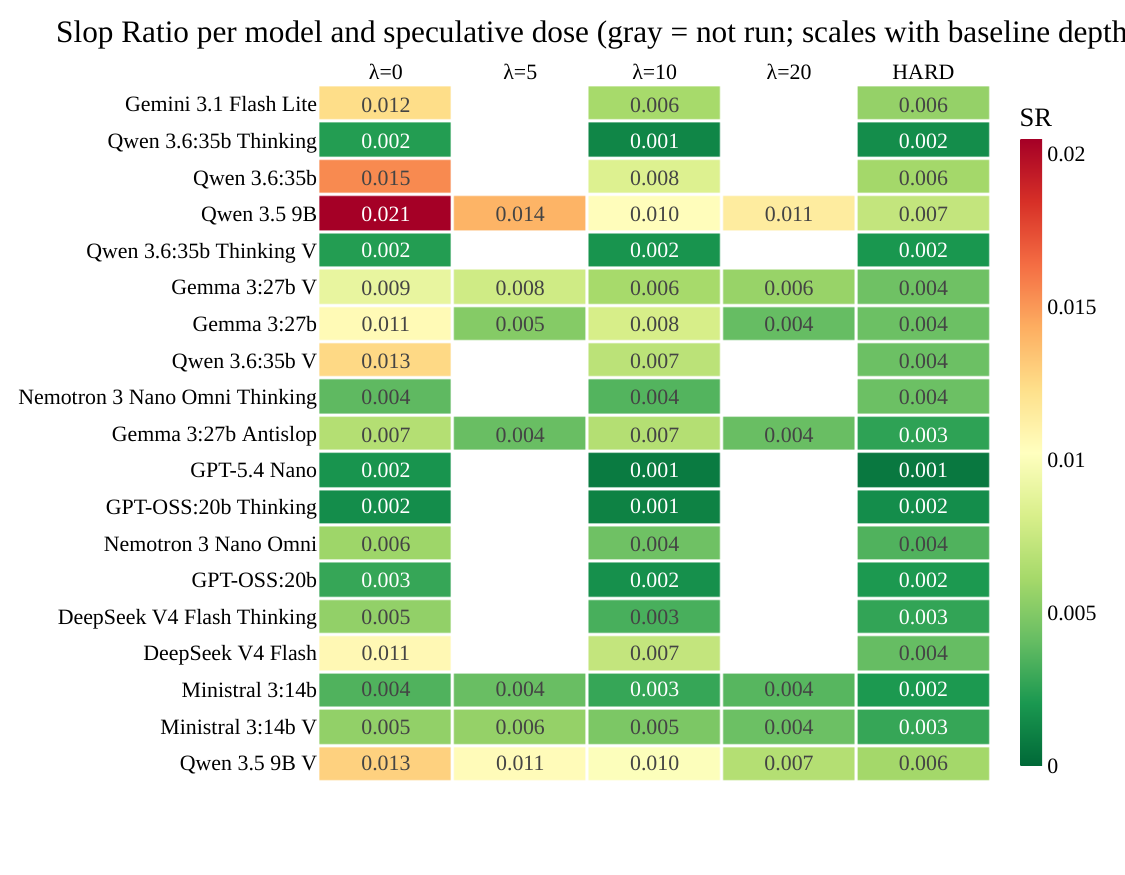}
\caption{Slop Ratio per model and speculative dose. Rows: the 19 models (Llama 3.3 70B excluded, sorted by baseline SR, deepest slop first). Columns: $\lambda = 0$, 5, 10, 20, and HARD; cell values printed, gray cells = dose not run for that model ($\lambda = 5$ and 20 cover the seven-model subset). The reduction grows with row depth and is monotone left to right in every model.}
\label{fig:dose_per_model}
\end{figure}

\section{Reasoning mode: thinking on versus off (RQ6)}
\label{app:B}

This appendix gives the full per-family breakdown behind the summary in Section~\ref{sec:reasoning}.
Four model families ship both a base and a thinking variant, which allows a controlled comparison at baseline.
Table~\ref{tab:B1} reports the same four metrics as Table~\ref{tab:C1}.

\begin{table}[t]
\centering
\caption{Baseline diversity metrics by reasoning mode. $\Delta$SR is thinking-on minus thinking-off; positive means thinking increases slop.}
\label{tab:B1}
\footnotesize
\begin{tabular}{lccccccccc}
\toprule
Model family & OFF SR & ON SR & $\Delta$SR & OFF $C_5$ & ON $C_5$ & OFF $H$ & ON $H$ & OFF $V$ & ON $V$ \\
\midrule
DeepSeek V4 Flash & 0.032 & 0.033 & $+0.001$ & 0.337 & 0.315 & 5.79 & 5.68 & 93 & 78 \\
GPT-OSS 20B & 0.034 & 0.035 & $+0.001$ & 0.393 & 0.385 & 5.16 & 5.31 & 54 & 64 \\
Nemotron 3 Nano Omni & 0.035 & 0.057 & $+0.022$ & 0.341 & 0.398 & 5.60 & 5.35 & 82 & 80 \\
Qwen 3.6 35B & 0.151 & 0.177 & $+0.026$ & 0.678 & 0.667 & 4.00 & 3.78 & 58 & 46 \\
\bottomrule
\end{tabular}
\end{table}

\section{Input modality: text versus vision (RQ2)}
\label{app:C}

This appendix gives the full per-family breakdown behind the summary in Section~\ref{sec:modality}.
Five model families ship both a text (chat) and a vision variant, which allows a controlled comparison at baseline.
Table~\ref{tab:C1} reports SR, top-5 concentration ($C_5$), Shannon entropy ($H$), and unique names ($V$) for each variant.

\begin{table}[t]
\centering
\caption{Baseline diversity metrics by input modality. SR is the Slop Ratio (lower is better). $C_5$ is top-5 concentration (lower is better). $H$ is Shannon entropy and $V$ is the number of unique names (higher is better for both). $\Delta$SR is vision minus text; negative means vision reduces slop.}
\label{tab:C1}
\footnotesize
\begin{tabular}{lccccccccc}
\toprule
Model family & Text SR & Vis SR & $\Delta$SR & Text $C_5$ & Vis $C_5$ & Text $H$ & Vis $H$ & Text $V$ & Vis $V$ \\
\midrule
Gemma 3 27B & 0.091 & 0.092 & $+0.001$ & 0.557 & 0.515 & 4.87 & 4.92 & 94 & 87 \\
Ministral 3 14B & 0.032 & 0.026 & $-0.006$ & 0.353 & 0.338 & 5.85 & 6.03 & 110 & 118 \\
Qwen 3.5 9B & 0.135 & 0.014 & $-0.121$ & 0.607 & 0.247 & 4.08 & 5.72 & 47 & 63 \\
Qwen 3.6 35B & 0.151 & 0.065 & $-0.086$ & 0.678 & 0.466 & 4.00 & 5.12 & 58 & 76 \\
Qwen 3.6 35B Thinking & 0.177 & 0.102 & $-0.075$ & 0.667 & 0.548 & 3.78 & 4.51 & 46 & 59 \\
\bottomrule
\end{tabular}
\end{table}

\section{Top names by language, per model}
\label{app:E}

This appendix gives the per-model breakdown behind the language analysis of Section~\ref{sec:dialects} (Table~\ref{tab:E1}).
For each chat and vision model, the two most frequent proper names at baseline are listed per language, with raw character counts.
Names are compared after accent-insensitive normalization, and each displayed form is the most frequent surface variant (\emph{El\'ias}/\emph{Elias} count together).
The table shows that some fingerprints travel across languages (Gemini Flash Lite's \emph{Elias}, Ministral's \emph{Elias}/\emph{Lena} pair) while others are locked to one language (Gemma's \emph{Tiber} in English).

\begin{table}[t]
\centering
\caption{Two most frequent proper names per model and language at baseline, with raw counts (200 stories per model; languages split 40/20/20/20\%). \emph{Elias} and \emph{El\'ias} are counted as one name; the displayed form is the most frequent variant.}
\label{tab:E1}
\scriptsize
\resizebox{\textwidth}{!}{%
\begin{tabular}{lp{3.6cm}p{2.6cm}p{2.6cm}p{2.6cm}}
\toprule
Model & English & Spanish & French & German \\
\midrule
Gemini 3.1 Flash Lite (text) & Elias (36), Elara (20) & Elian (13), El\'ias (9) & Elias (10), Arthur (7) & Elias (20), Elara (3) \\
Qwen 3.6:35b Thinking (text) & Elara (43), Elias (6) & Elara (7), Clara (4) & \'Elias (5), L\'ea (5) & Elara (7), Leo (4) \\
Qwen 3.6:35b (text) & Elara (45), Barnaby (17) & Elara (7), El\'ias (6) & Elara (10), L\'eo (10) & Elara (12), Leo (5) \\
Qwen 3.5 9B (text) & Elara (34), Barnaby (10) & Leo (4), Alex (3) & Elara (9), L\'eo (5) & Elias (5), Elara (4) \\
Qwen 3.6:35b Thinking (vis) & Elara (20), Leo (19) & Leo (9), Lucas (3) & L\'eo (9), Barnab\'e (2) & Leo (7), Elara (5) \\
Gemma 3:27b (vis) & Tiber (70), Leo (10) & Elara (7), Leo (7) & L\'eo (12), Barnab\'e (3) & Elara (8), Elias (4) \\
Gemma 3:27b (text) & Tiber (67), Elara (15) & Silas (12), El\'ias (7) & Elara (7), Elias (5) & Elara (14), Elias (7) \\
Qwen 3.6:35b (vis) & Elara (18), Leo (18) & Leo (5), L\'azaro (3) & L\'eo (6), Lila (4) & Leo (6), Elara (4) \\
Nemotron 3 Nano Omni Thinking & Maya (34), Lina (7) & Ana (8), L\'ia (2) & L\'ea (7), Elara (2) & Lena (8), Auge (3) \\
Gemma 3:27b Antislop (text) & Tiber (47), Flora (15) & Mateo (6), Elena (5) & Elias (5), L\'ea (2) & Elias (7), Lena (3) \\
GPT-5.4 Nano (text) & Mira (13), Mara (6) & L\'azaro (2), Sof\'ia (2) & L\'eo (3), Lila (3) & Mira (6), Liora (2) \\
GPT-OSS:20b Thinking (text) & Mara (16), Maya (11) & Ana (4), Mateo (2) & L\'eo (7), L'ali\'en\'e (2) & Lina (6), Max (3) \\
Nemotron 3 Nano Omni (text) & Elara (25), Lila (10) & Clara (5), L\'ia (5) & \'Elodie (7), L\'ea (5) & Lena (4), Elias (2) \\
GPT-OSS:20b (text) & Mira (9), Mara (8) & Ana (4), L\'ia (2) & L\'eo (13), Mira (2) & Lila (3), Lina (3) \\
DeepSeek V4 Flash Thinking & Leo (10), Elara (6) & Marcos (3), Leo (2) & L\'eo (7), Barbe-Rouge (2) & Lena (4), Elias (3) \\
DeepSeek V4 Flash (text) & Elara (15), Leo (13) & Leo (6), Marcos (5) & Barbe-Rouge (3), L\'eo (3) & Lena (5), Ph\"onix (3) \\
Ministral 3:14b (text) & Elias (12), Lena (9) & Luna (11), Clara (4) & Elias (7), L\'ea (4) & Elias (10), Lina (9) \\
Llama 3.3:70b (text) & Cowboy (2), I (2) & Cara (2) & --- & Abenteurer (4), Auge (1) \\
Ministral 3:14b (vis) & Lila (11), Liam (8) & Luna (8), Leo (3) & L\'eo (10), L\'ea (3) & Lina (16), Lena (5) \\
Qwen 3.5 9B (vis) & Leo (7), Luna (3) & Ana (1), Apple (1) & L\'eon (4), L\'ea (2) & Alien (1), Elara (1) \\
\bottomrule
\end{tabular}%
}
\end{table}

\section{Sentence-opener concentration: proof of concept}
\label{app:F}

This appendix documents the proof of concept behind the measured support for the transfer hypothesis of Section~\ref{sec:beyond-names}.
We extract the first four words of every sentence (story openers and mid-story sentences separately), abstract proper names to \texttt{$<$NAME$>$}, filter pipeline meta-prefaces, and pool all generations.
The first author reviewed the top 20 candidate openers per language and bucket for English and Spanish (80 candidates): 60 were confirmed as slop, 16 were rated uncertain and excluded from the rate, and 4 were legitimate (a confirmed-slop rate of 94\%).
Table~\ref{tab:opener_poc} lists the two most frequent openers per language and bucket with their share and the number of models whose pooled output contains them; recurrence across independent models is high even where the exact share is low.

\begin{table}[t]
\centering
\footnotesize
\setlength{\tabcolsep}{4pt}
\caption{Sentence-opener concentration, proof of concept (all generations pooled; four-gram initials, proper names abstracted to \texttt{$<$NAME$>$}; meta-prefaces filtered). Share = percentage of sentences in the cell starting with the opener; models = number of models whose pooled output contains it (20 per language).}
\label{tab:opener_poc}
\resizebox{\textwidth}{!}{%
\begin{tabular}{lllrrlrr}
\toprule
Lang. & Bucket & Dominant opener & Share (\%) & Models & Second opener & Share (\%) & Models \\
\midrule
en & story first & \texttt{in the heart of} & 1.5 & 13/20 & \texttt{in a quiet village} & 1.1 & 11/20 \\
en & mid-story & \texttt{with a deep breath} & 0.5 & 17/20 & \texttt{with a flick of} & 0.1 & 16/20 \\
es & story first & \texttt{en un mundo donde} & 1.4 & 5/20 & \texttt{en un bosque encantado} & 1.3 & 10/20 \\
es & mid-story & \texttt{al llegar a la} & 0.4 & 20/20 & \texttt{de repente una gota} & 0.2 & 16/20 \\
fr & story first & \texttt{dans un monde où} & 2.6 & 8/20 & \texttt{une simple goutte d'eau} & 2.0 & 6/20 \\
fr & mid-story & \texttt{il comprit alors que} & 0.2 & 19/20 & \texttt{après des jours de} & 0.2 & 17/20 \\
de & story first & \texttt{in einer welt in} & 1.3 & 10/20 & \texttt{auf einer einsamen insel} & 0.9 & 14/20 \\
de & mid-story & \texttt{mit einem tiefen atemzug} & 0.2 & 13/20 & \texttt{es war kein $<$NAME$>$} & 0.1 & 10/20 \\
\bottomrule
\end{tabular}}
\end{table}

\end{document}